\documentclass[12pt,letterpaper]{article}
\usepackage[a4paper, total={7in, 10in}]{geometry}

\usepackage{graphicx}
\usepackage{helvet}
\usepackage{authblk}
\usepackage{hyperref}
\usepackage{amsmath} 
\usepackage{amssymb} 
\usepackage{orcidlink} 
\usepackage[super,comma,sort&compress]  
   {natbib}
\usepackage[title]{appendix}%

\usepackage[utf8]{inputenc} 
\usepackage[T1]{fontenc}    
\usepackage{url}            
\usepackage{booktabs}       
\usepackage{amsfonts}       
\usepackage{nicefrac}       
\usepackage{microtype}      
\usepackage{xcolor}         
\usepackage{multirow}
\usepackage{geometry}
\usepackage{threeparttable}
\usepackage{rotating}
\usepackage{graphicx}
\usepackage{amsmath}
\usepackage{subcaption}
\usepackage{placeins}
\usepackage[utf8]{inputenc}
\usepackage{booktabs}
\usepackage{siunitx}
\usepackage{caption}
\usepackage{tabularx}
\usepackage{listings}
\usepackage{pifont}
\usepackage{array}
\usepackage{makecell}
\usepackage{amssymb}
\theadset{\bfseries}

\makeatletter
\renewcommand{\maketitle}{\bgroup\setlength{\parindent}{0pt}
\begin{flushleft}
  \textbf{\@title}
  
  \@author
\end{flushleft}\egroup}
\makeatother

\title{SpiroLLM: Finetuning Pretrained LLMs to Understand Spirogram Time Series with Clinical Validation in COPD Reporting}
\date{}

\author[1,2,8\orcidlink{0009-0007-9505-3699}]{Shuhao Mei}
\author[2\orcidlink{0009-0003-8345-2212}]{Yongchao Long}
\author[1]{Xiaoyu Xiao}
\author[3]{Shan Cao}
\author[4]{Xiaobo Han}
\author[5]{Shijia Geng}
\author[1,6,*]{Jinbo Sun}
\author[2,7,*]{Yuxi Zhou}
\author[8,9,*,\orcidlink{0000-0001-7521-5127}]{Shenda Hong}

\affil[1]{Guangzhou Institute of Technology, Xidian University, Xi'an, China}
\affil[2]{Department of Computer Science, Tianjin University of Technology, Tianjin, China}
\affil[3]{Department of Respiratory, The Second Hospital of Tianjin Medical University, China}
\affil[4]{College of Pulmonary and Critical Care Medicine, Chinese PLA General Hospital, Beijing, China}
\affil[5]{HeartVoice Medical Technology, Hefei, China}
\affil[6]{School of Life Science and Technology, Xidian University, Xi'an, China}
\affil[7]{DCST, BNRist, RIIT, Institute of Internet Industry, Tsinghua University, Beijing, China}
\affil[8]{National Institute of Health Data Science, Peking University, Beijing, China}
\affil[9]{Institute for Artificial Intelligence, Peking University, Beijing, China}

\affil[*]{Correspondence: hongshenda@pku.edu.cn, joy\_yuxi@pku.edu.cn, sunjb@xidian.edu.cn}

\begin{document}

\maketitle

\section*{ABSTRACT}
Chronic Obstructive Pulmonary Disease (COPD), a major chronic respiratory disease with persistent airflow limitation, is a leading global cause of disability and mortality. Respiratory spirogram time series, routinely collected during pulmonary function tests (PFTs), play a critical role in the early detection of repsiratory diseases and in monitoring lung function over time.
However, most current AI models for COPD diagnosis are limited to outputting classification results without providing a rationale for their diagnostic process, while current Large Language Models (LLMs) cannot understand spirograms yet, which severely limits their clinical trust and adoption. To tackle this challenge, we leverage a cohort of 234,028 individuals from the UK Biobank (UKB) to propose SpiroLLM, the first multimodal large language model that can understand spirogram. The model extracts morphological features from respiratory curves via a SpiroEncoder and aligns them with PFT numerical values in a unified latent space using a SpiroProjector, ultimately empowering a large language model to generate a comprehensive diagnostic report. 
Experimental results confirm that SpiroLLM achieved a diagnostic AUROC of 0.8977 (95\% CI: 0.88-0.91). In a robustness test with missing core data, it maintained a 100\% valid response rate, far surpassing the 13.4\% of a text-only model and showcasing the superiority of its multimodal design. This work demonstrates the substantial potential of deeply fusing physiological signals with large language models, establishing a new paradigm for the next generation of interpretable and reliable clinical decision support tools.

\section*{KEYWORDS}
Chronic Obstructive Pulmonary Disease, Large Language Models, Multimodal Fusion, Automated Report Generation

\section*{AUTHOR SUMMARY}
Chronic Obstructive Pulmonary Disease (COPD) is a leading cause of death and disability worldwide, yet diagnosing it remains a significant challenge. Accurate diagnosis typically requires highly trained specialists to interpret complex charts from breathing tests—a process that is time-consuming and often unavailable in areas with limited medical resources. While artificial intelligence has shown promise in medicine, most existing tools only provide simple "yes or no" classifications without explaining the reasoning behind the diagnosis, which limits their usefulness and trustworthiness for doctors.
In this study, we developed a new system called SpiroLLM to bridge this gap. Unlike traditional tools, our approach uses advanced technology to "look" at the shape of a patient’s breathing curves and combine this visual information with standard test numbers. The system then automatically writes a detailed, easy-to-understand clinical report, much like a human expert would. We tested our model using a large dataset from the UK Biobank and found it to be highly accurate and reliable, even when some patient data was missing. This work represents a step forward in digital health, offering a transparent assistant for clinicians that could make expert-level COPD diagnosis more accessible and consistent globally.

\section*{INTRODUCTION}

Chronic Obstructive Pulmonary Disease (COPD), a major chronic respiratory disease characterized by persistent airflow limitation, is one of the leading causes of disability and mortality worldwide\cite{Venkatesan2025}. In the clinical diagnosis and management of COPD, the Pulmonary Function Test (PFT) and its corresponding spirogram curve are indispensable. They not only represent the gold standard for diagnosis but also serve as a crucial basis for assessing disease severity, monitoring progression, and guiding treatment strategies\cite{agusti2022global}. However, the accurate interpretation of spirogram curves and the subsequent drafting of a standardized yet personalized diagnostic report are time-consuming, labor-intensive processes that are highly dependent on the specialized knowledge and long-term experience of clinicians. This reliance on expert resources is particularly pronounced in regions with limited medical access, creating a significant bottleneck in improving the efficiency and standardization of COPD diagnosis\cite{stanojevic2022ers}.

To address this challenge, researchers have begun exploring the use of Artificial Intelligence (AI) to automate diagnostics\cite{das2023collaboration}. In our prior work, we developed DeepSpiro \cite{mei2025deep}, a model that demonstrated the feasibility of using deep learning to identify COPD-related features directly from spirogram curves. However, this and other early deep learning models were limited by their "black-box" nature, outputting only simple classification labels. Their inability to provide a rationale for their conclusions has hindered their clinical adoption and trust. More recently, the advent of Large Language Models (LLMs) has shown great promise in addressing this interpretability issue, with their ability to generate logically coherent medical texts that emulate the style of human experts \cite{liu2025application}. Nevertheless, applying LLMs to generate diagnostic reports directly from raw pulmonary function data still faces three core challenges:

\begin{figure*}[]
\centerline{\includegraphics[width=1\textwidth]{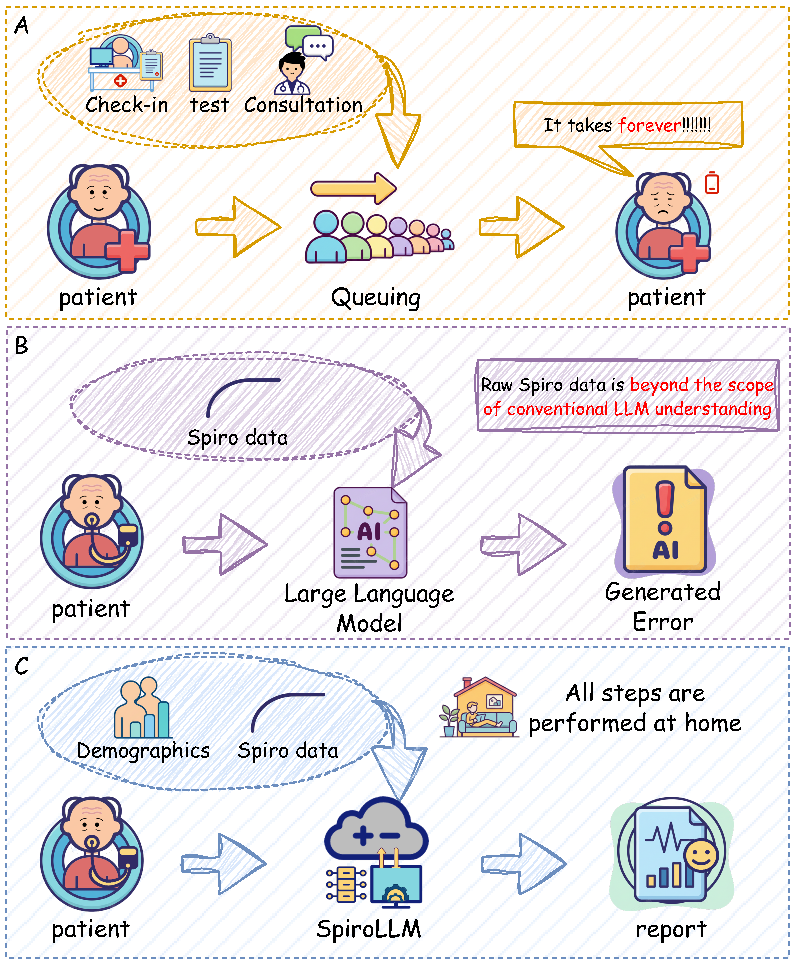}}
\caption{This figure compares three workflows for pulmonary function assessment: the traditional clinical model (A), which relies on cumbersome in-clinic testing; the traditional large language model (B), which cannot understand raw physiological signals; and our proposed SpiroLLM framework (C), which supports at-home self-testing and instant generation of professional reports, significantly improving efficiency.}
\label{fig:overview}
\end{figure*}

\begin{itemize}
\item A fundamental disconnect exists in current approaches. On one hand, vision-based or sequential models can process spirogram curves but cannot generate comprehensive reports. On the other hand, LLMs excel at processing textualized PFT numerical data but cannot directly "see" and interpret the rich morphological information embedded in the waveforms. A unified, end-to-end framework that seamlessly integrates both modalities is currently lacking.
\item Training a reliable report generation model requires a massive volume of high-quality, expert-authored reports as supervision signals. In clinical practice, it is infeasible to have specialists manually annotate tens of thousands of samples, creating a critical bottleneck at the data level.
\item The evaluation of current generative models largely relies on conventional text-similarity metrics (e.g., ROUGE, BLEU). These metrics fail to effectively measure performance along critical dimensions such as medical factual accuracy, logical coherence, and clinical safety, and thus do not reflect the true clinical utility of the models.
\end{itemize}

To address the aforementioned challenges, we leveraged the authoritative, large-scale UK Biobank (UKB) to develop and validate SpiroLLM—a framework for COPD diagnostic report generation based on multimodal fusion and large language models (as shown in Figure \ref{fig:overview}). The main contributions of this study are as follows:

\begin{itemize}
    \item Building on our prior work in spirogram feature analysis, we are the first to design and implement SpiroLLM, which seamlessly integrates a specialized SpiroEncoder (for encoding spirogram curves) with an LLM via a lightweight alignment module, the SpiroProjector. This architecture achieves, for the first time, a deep fusion of visual features from time-series waveforms and textual PFT metrics, enabling the model to perform end-to-end diagnostic report generation.
    \item To alleviate the scarcity of annotated data, we developed a semi-automated report generation pipeline. This pipeline combines a vision-language model, a quantitative metric calculation module, and a Retrieval-Augmented Generation mechanism based on GOLD guidelines. This process significantly reduces the cost and burden of manual annotation while ensuring the authoritativeness of the diagnostic logic.
    \item We adopted an "LLM-as-a-Judge" approach to establish an evaluation framework spanning six clinical dimensions, including factual accuracy, logical consistency, and completeness. Furthermore, through meticulously designed input masking experiments, we quantitatively verify the superior robustness of our multimodal approach compared to single-modality methods and confirm the independent diagnostic contribution of visual features.
\end{itemize}

SpiroLLM is not only a technical innovation but also poised to become a powerful assistant for clinicians. By enhancing the efficiency and consistency of diagnostic report writing, it promises to ultimately improve patient care experiences and long-term health management.

\section*{RESULTS}
\subsection*{Method Overview}
\begin{figure*}[h]
\centerline{\includegraphics[width=1\textwidth]{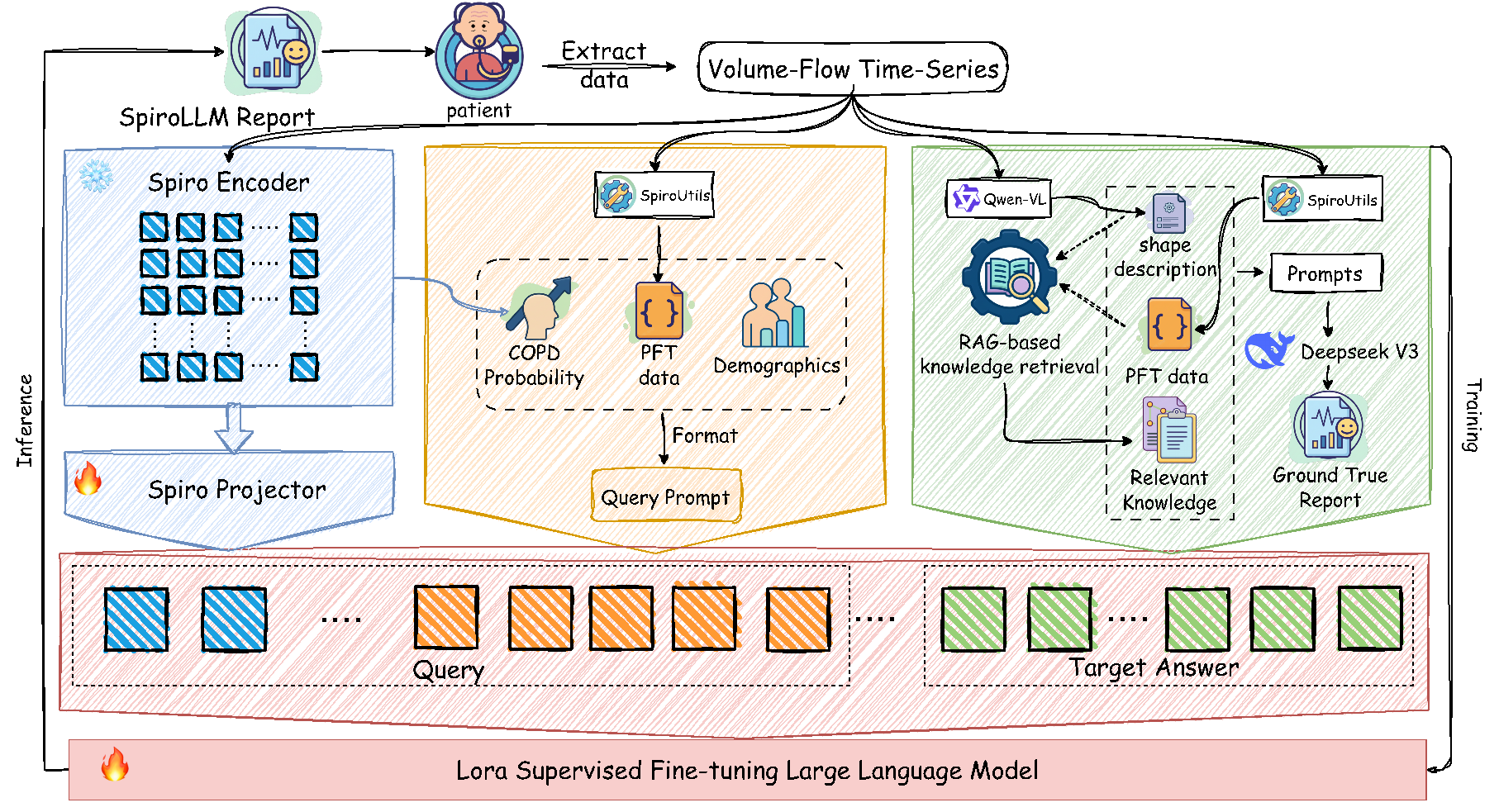}}
\caption{A schematic diagram of the overall architecture of the SpiroLLM framework. The figure illustrates the complete end-to-end process, from raw pulmonary function test time-series data to the generation of a professional diagnostic report. The blue section represents the SpiroEncoder module, which extracts high-level features from the spirometry curves and performs cross-modal alignment with the large language model via the SpiroProjector. The yellow section is the Query Prompt construction module, which integrates the COPD probability output by the SpiroEncoder, key PFT parameters extracted by SpiroUtils, and the patient's demographic information to form the model's input prompt. The green section represents the gold-standard report generation process. This process begins by using the Qwen-VL model to generate morphological descriptions from the pulmonary function curve images, it then incorporates the PFT values extracted by SpiroUtils and introduces relevant domain knowledge through a RAG-based knowledge base system. Finally, all this information is integrated by the DeepSeek V3 model to generate a high-quality, standardized diagnostic report, which serves as the training target output for SpiroLLM.}
\label{fig:framework}
\end{figure*}

Our methodology centers on the development of 
SpiroLLM, a multimodal large language model that automatically generates clinical reports for COPD from patient data. As illustrated in our framework (Figure \ref{fig:framework}), the process begins with the SpiroEncoder, a hybrid CNN-BiLSTM network, which extracts deep feature embeddings from raw spirometry time-series data. To bridge the modality gap between these numerical features and the text-based domain of the LLM, a lightweight MLP called the SpiroProjector aligns the signal features with the LLM's embedding space. These projected features are then combined with the patient's demographic information to create a multimodal prompt that is fed into the core language model. A key contribution of our work is the generation of high-quality "gold-standard" reports for supervised fine-tuning. We designed a semi-automated pipeline that synthesizes three crucial pieces of information: (1) qualitative morphological descriptions of the spirometry curve generated by a visual language model (Qwen-VL), (2) quantitative physiological metrics calculated by our SpiroUtils tool, and (3) relevant clinical knowledge retrieved from a GOLD standard knowledge base using Retrieval-Augmented Generation (RAG). These components are integrated by the DeepSeek-V3 model to produce a comprehensive target report. The entire SpiroLLM is then trained efficiently using the LoRA parameter-efficient fine-tuning strategy. Finally, we evaluate the model's performance using an "LLM-as-a-Judge" approach, where an independent LLM assesses both the clinical quality of the generated reports and their diagnostic accuracy (AUROC, AUPRC, F1-Score).

\subsection*{Diagnostic Performance Analysis}
We evaluated the diagnostic performance of SpiroLLM against the comprehensive set of baselines detailed in the Methods. Table \ref{tab:diagnostic_accuracy} presents the comprehensive diagnostic performance comparison. Statistical significance was assessed using Bootstrap resampling ($n=10000$).

\begin{sidewaystable}[p] 
\centering
\begin{threeparttable}
    \caption{\textbf{Comparison of performance among different methods.} The table reports the Mean (95\% Confidence Interval). The proposed SpiroLLM demonstrates superior performance, particularly in Sensitivity, compared to all baselines.}
    \label{tab:diagnostic_accuracy}
    
    \renewcommand{\arraystretch}{1.4}
    \setlength{\tabcolsep}{5pt}
    \footnotesize

    \begin{tabular}{llccccc}
    \toprule
    \textbf{Category} & \textbf{Model} & \textbf{AUROC} & \textbf{AUPRC} & \textbf{Sensitivity} & \textbf{Specificity} & \textbf{F1 Score} \\ 
    \midrule

    \multirow{3}{*}{\shortstack[l]{Demographic \& \\ Clinical Baselines}} 
     & XGBoost-Demo \cite{chen2016xgboost} & 0.7288 (0.71-0.75) & 0.6497 (0.61-0.69) & 0.7082 (0.68-0.74) & 0.6456 (0.62-0.67) & 0.6597 (0.63-0.68) \\
     & XGBoost-Anthro \cite{chen2016xgboost} & 0.7368 (0.71-0.76) & 0.6816 (0.65-0.71) & 0.7703 (0.74-0.80) & 0.5953 (0.57-0.62) & 0.6783 (0.65-0.70) \\
     & XGBoost-Clinical \cite{chen2016xgboost} & 0.8573 (0.84-0.87) & 0.8331 (0.81-0.86) & 0.7377 (0.71-0.77) & 0.8292 (0.81-0.85) & 0.7569 (0.73-0.78) \\ 
    \midrule

    \multirow{5}{*}{\shortstack[l]{Time-Series \\ Baselines}} 
     & Informer \cite{zhou2021informer} & 0.7687 (0.75-0.79) & 0.7098 (0.68-0.74) & 0.7351 (0.70-0.77) & 0.6800 (0.65-0.71) & 0.6898 (0.66-0.71) \\
     & PatchTST \cite{Yuqietal-2023-PatchTST} & 0.8073 (0.79-0.83) & 0.7810 (0.75-0.81) & 0.7389 (0.71-0.77) & 0.7378 (0.71-0.76) & 0.7161 (0.69-0.74) \\
     & TimesNet \cite{wu2022timesnet} & 0.8270 (0.81-0.85) & 0.8035 (0.78-0.83) & 0.7582 (0.73-0.79) & 0.7643 (0.74-0.79) & 0.7397 (0.72-0.76) \\
     & SpiroEncoder \cite{mei2025deep} & 0.8249 (0.81-0.84) & 0.8041 (0.78-0.83) & 0.7474 (0.72-0.78) & 0.7467 (0.72-0.77) & 0.7253 (0.70-0.75) \\
     & ResNet18 \cite{cosentino2023inference} & 0.8355 (0.82-0.85) & 0.8105 (0.78-0.84) & 0.7401 (0.71-0.77) & 0.7869 (0.76-0.81) & 0.7386 (0.71-0.76) \\
    \midrule

    \multirow{1}{*}{\shortstack[l]{Multimodal Fusion \\ Baseline}} 
     & DeepSpiro \cite{mei2025deep} & 0.8625 (0.85-0.88) & 0.8420 (0.82-0.87) & 0.7838 (0.76-0.81) & 0.7860 (0.76-0.81) & 0.7651 (0.74-0.79) \\ 
    \midrule

    \multirow{2}{*}{\shortstack[l]{Proposed \\ Methods}} 
     & SpiroLLM (PFT-only) & 0.8963$^{\dagger\ddagger}$ (0.88-0.91) & 0.9009$^{\dagger\ddagger}$ (0.89-0.92) & 0.7764 (0.75-0.80) & \textbf{0.9960}$^{\dagger\ddagger}$ (0.99-1.00) & 0.8717$^{\dagger\ddagger}$ (0.85-0.89) \\
     & \textbf{SpiroLLM} & \textbf{0.8977}$^{\dagger\ddagger}$ (0.88-0.91) & \textbf{0.9046}$^{\dagger\ddagger}$ (0.89-0.92) & \textbf{0.8165}$^{\dagger\ddagger*}$ (0.79-0.84) & 0.9931$^{\dagger\ddagger}$ (0.99-1.00) & \textbf{0.8947}$^{\dagger\ddagger*}$ (0.88-0.91) \\ 
    \bottomrule
    \end{tabular}

    \begin{tablenotes}
        \small
        \item \textit{Notes:} 
        \item $^\dagger$ Significant improvement over \textbf{DeepSpiro} ($p < 0.001$).
        \item $^\ddagger$ Significant improvement over \textbf{XGBoost-Clinical} ($p < 0.001$).
        \item $^*$ Significant improvement over \textbf{SpiroLLM (PFT-only)} ($p < 0.001$).
        \item \textbf{Model Inputs Definition:}
        \item $\bullet$ \textbf{XGBoost-Demo}: Age, Sex, Smoking Status.
        \item $\bullet$ \textbf{XGBoost-Anthro}: XGBoost-Demo features + Height, BMI.
        \item $\bullet$ \textbf{XGBoost-Clinical}: XGBoost-Anthro features + PFT metrics derived from SpiroUtils (FEV1, FVC, etc.).
        \item $\bullet$ \textbf{Time-Series Baselines}: Raw flow-volume curve data.
        \item $\bullet$ \textbf{DeepSpiro}: Raw flow-volume curve data + Age, Sex, Smoking, and FEV1/FVC ratio.
        \item $\bullet$ \textbf{SpiroLLM (PFT-only)}: Basic Info (Age, Sex, Height, Smoking) + Comprehensive PFT Results (FEV1, FVC, FEV1/FVC, PEF, FEF25-75, including Measured, Predicted, LLN, and Z-scores).
        \item $\bullet$ \textbf{SpiroLLM}: Fusion of SpiroLLM (PFT-only) inputs + Raw flow-volume curve data.
    \end{tablenotes}
\end{threeparttable}
\end{sidewaystable}

Consistent with clinical expectations, the progressive inclusion of comprehensive spirometry metrics effectively enhanced baseline performance. The XGBoost-Clinical model established a robust baseline, achieving an AUROC of 0.8573. SpiroLLM further elevated the AUROC to 0.8977, significantly outperforming this clinical baseline model ($p < 0.001$). This result provides empirical evidence that raw flow-volume curves encapsulate subtle morphological patterns. These patterns, effectively captured by our visual encoder, offer complementary information to discrete pulmonary function test (PFT) metrics that cannot be fully characterized by numerical data alone.

Furthermore, SpiroLLM outperformed all unimodal time-series encoders (e.g., TimesNet, AUROC 0.8270) as well as the previously proposed multimodal benchmark, DeepSpiro (AUROC 0.8625, $p < 0.001$). This underscores the architectural advantage of the proposed method: by aligning visual representations with the semantic reasoning capabilities of a Large Language Model (LLM), it achieves superior diagnostic discrimination compared to the traditional feature concatenation strategies employed in prior studies.

A critical comparison lies between the text-only variant (SpiroLLM PFT-only) and the complete SpiroLLM framework. Although their overall discriminative ability (AUROC) was comparable (0.8963 vs. 0.8977), the incorporation of raw flow-volume curves yielded a statistically significant improvement in sensitivity (81.65\% vs. 77.64\%, $p < 0.001$). This absolute improvement of approximately 4.0\% highlights the distinct clinical value of the visual modality: it enables the identification of "hard-to-diagnose" positive cases that present ambiguously in numerical metrics, thereby effectively reducing the rate of missed diagnoses in screening scenarios.

\subsection*{Visual Interpretability and Morphological Analysis}
To investigate whether the observed performance enhancements—particularly the increased sensitivity within the multimodal framework—stem from clinically meaningful feature extraction rather than data artifacts, we visualized the attention distribution of the visual encoder (SpiroEncoder). By extracting activation maps from the final layer of the encoder and mapping them onto the raw flow-volume curves, we were able to interpret the specific regions prioritized by the model.

\begin{figure*}[h]
\centerline{\includegraphics[width=1\textwidth]{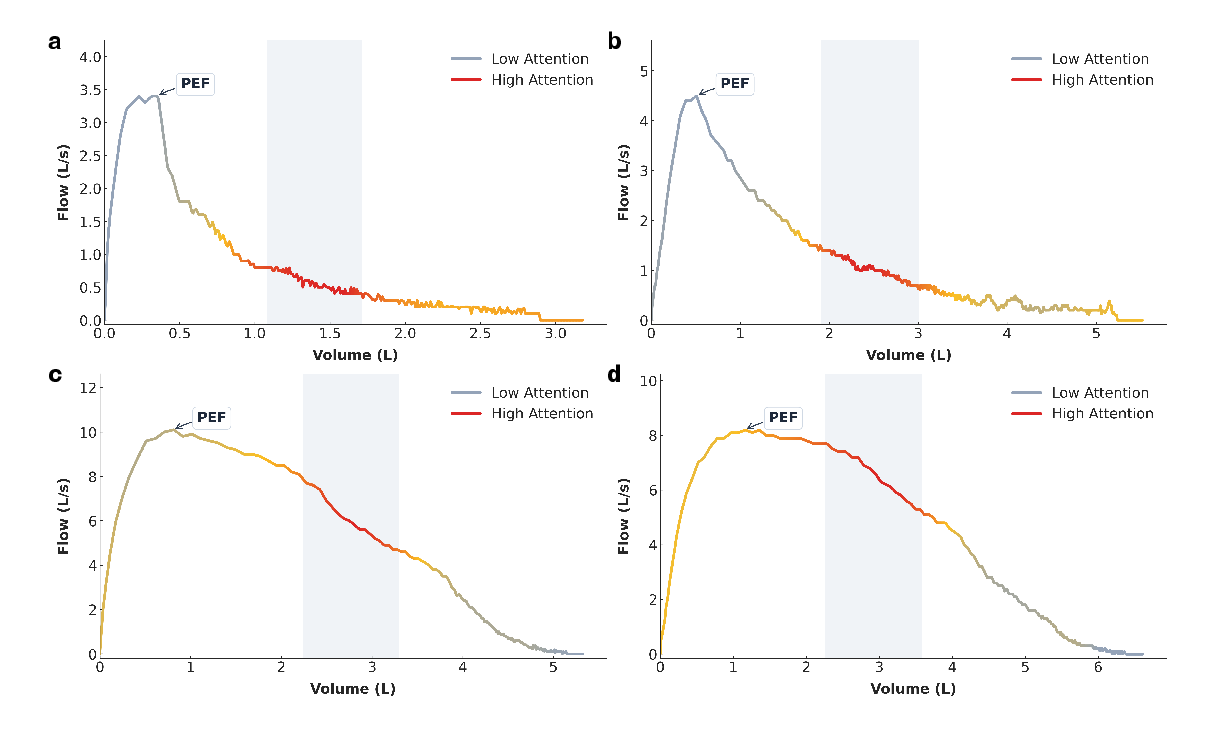}}
\caption{Interpretable visualization of the SpiroEncoder. Flow-volume curves are color-coded based on attention weights extracted from the final layer of the encoder (red indicates high attention; gray indicates low attention). (a-b) COPD cases: The model exhibits high attention toward the concave descending limb, successfully identifying the characteristic "scooped" pattern associated with airway obstruction. (c-d) Normal cases: Attention remains focused on the descending limb, validating the linear or convex profile associated with normal lung function. This confirms that the model explicitly learns to analyze morphological features indicative of small airway obstruction.}
\label{fig:SpiroEncoder_results}
\end{figure*}

As illustrated in Figure \ref{fig:SpiroEncoder_results}, the generated heatmaps reveal a high degree of alignment between the features learned by the model and clinical expert intuition. Across all subjects, the model consistently concentrated regions of high attention (indicated in red) along the descending limb of the expiratory curve. In clinical physiology, this segment corresponds to variations in small airway airflow (FEF25–75\%) and represents the primary site of airflow obstruction in Chronic Obstructive Pulmonary Disease (COPD). Specifically, in positive cases (Figures \ref{fig:SpiroEncoder_results}a and \ref{fig:SpiroEncoder_results}b), the model exhibited significant focus on the characteristic concavity or "scooped" pattern appearing after peak flow. Conversely, in normal cases (Figures \ref{fig:SpiroEncoder_results}c and \ref{fig:SpiroEncoder_results}d), attention was directed toward the relatively linear or convex trajectory of the descending limb.

These visualization results empirically demonstrate that the visual encoder serves as a critical component capable of autonomously learning to identify characteristic morphological landmarks of airflow limitation, thereby providing a transparent and interpretable basis for the model's diagnostic decisions.

\subsection*{Quality of Generated Reports}
In terms of the quality of the generated reports, model fine-tuning also brought significant improvements. As shown in Table \ref{tab:quality_of_generated_report}, SpiroLLM and its variant, SpiroLLM-pftonly, scored significantly higher than the Llama3.1-8B baseline model across multiple evaluation dimensions, including factual accuracy, content completeness, and logical coherence. This fully validates the effectiveness of our proposed fine-tuning framework in guiding the model to generate specialized medical reports. SpiroLLM-pftonly, by leveraging complete textualized PFT values, is also able to generate high-quality diagnostic reports, with its performance being comparable to that of SpiroLLM.

\begin{table*}[h]
    \centering
    \small
    \renewcommand{\arraystretch}{1.5}

    \caption{Comparison of the quality of generated COPD diagnostic reports across different methods. Values are presented as mean (95\% confidence interval).}
    \label{tab:quality_of_generated_report}

    \begin{tabular*}{\textwidth}{
        l
        @{\extracolsep{\fill}}
        S[table-format=2.2]
        S[table-format=2.2]
        S[table-format=2.2]
        S[table-format=2.2]
        S[table-format=2.2]
        S[table-format=2.2]
    }
        \toprule
        \textbf{Method} &
        \textbf{\begin{tabular}{@{}c@{}}Factual \\ Accuracy\end{tabular}} &
        \textbf{\begin{tabular}{@{}c@{}}Completeness \\ \& Coverage\end{tabular}} &
        \textbf{\begin{tabular}{@{}c@{}}Logic \& \\ Evidence\end{tabular}} &
        \textbf{\begin{tabular}{@{}c@{}}Medical \\ Terminology\end{tabular}} &
        \textbf{\begin{tabular}{@{}c@{}}Medical \\ Safety\end{tabular}} &
        \textbf{\begin{tabular}{@{}c@{}}Curve \\ Description\end{tabular}} \\
        \midrule
        Llama3.1-8B\cite{grattafiori2024llama} & {\begin{tabular}{@{}c@{}}48.56 \\[-0.5ex] \tiny (47.56-49.57)\end{tabular}} & {\begin{tabular}{@{}c@{}}64.22 \\[-0.5ex] \tiny (63.65-64.79)\end{tabular}} & {\begin{tabular}{@{}c@{}}49.39 \\[-0.5ex] \tiny (48.35-50.42)\end{tabular}} & {\begin{tabular}{@{}c@{}}84.10 \\[-0.5ex] \tiny (83.55-84.67)\end{tabular}} & {\begin{tabular}{@{}c@{}}66.83 \\[-0.5ex] \tiny (65.41-68.26)\end{tabular}} & {\begin{tabular}{@{}c@{}}33.48 \\[-0.5ex] \tiny (32.42-34.54)\end{tabular}} \\
        
        \begin{tabular}{@{}c@{}}SpiroLLM \\[-0.5ex] \small (PFT-only)\end{tabular} & {\begin{tabular}{@{}c@{}}79.86 \\[-0.5ex] \tiny (78.72-80.91)\end{tabular}} & {\begin{tabular}{@{}c@{}}86.20 \\[-0.5ex] \tiny (85.42-86.92)\end{tabular}} & {\begin{tabular}{@{}c@{}}83.64 \\[-0.5ex] \tiny (82.46-84.74)\end{tabular}} & {\begin{tabular}{@{}c@{}}96.46 \\[-0.5ex] \tiny (96.05-96.85)\end{tabular}} & {\begin{tabular}{@{}c@{}}91.92 \\[-0.5ex] \tiny (91.01-92.77)\end{tabular}} & {\begin{tabular}{@{}c@{}}86.53 \\[-0.5ex] \tiny (85.58-87.44)\end{tabular}} \\
        
        SpiroLLM                              & {\begin{tabular}{@{}c@{}}78.36 \\[-0.5ex] \tiny (77.12-79.56)\end{tabular}} & {\begin{tabular}{@{}c@{}}86.39 \\[-0.5ex] \tiny (85.66-87.13)\end{tabular}} & {\begin{tabular}{@{}c@{}}81.63 \\[-0.5ex] \tiny (80.36-82.88)\end{tabular}} & {\begin{tabular}{@{}c@{}}95.62 \\[-0.5ex] \tiny (95.17-96.04)\end{tabular}} & {\begin{tabular}{@{}c@{}}89.03 \\[-0.5ex] \tiny (87.97-90.06)\end{tabular}} & {\begin{tabular}{@{}c@{}}85.76 \\[-0.5ex] \tiny (84.74-86.76)\end{tabular}} \\
        \bottomrule
    \end{tabular*}
\end{table*}

\subsection*{Robustness Test}
Based on the results above, it is evident that under ideal conditions where all relevent information is fully accessible, SpiroLLM-pftonly demonstrates strong competitiveness. However, a key question arises: Is this performance robust when essential inputs are missing? Specifically, can the model maintain its effectiveness without explicit access to PFT numerical values in the text? To explore this, we designed a robustness test to evaluate the model's generalization ability under conditions of missing information or environmental uncertainty.

\begin{table*}[h]
    \centering
    \small
    \renewcommand{\arraystretch}{1.5}

    \caption{Ablation study of SpiroLLM. We compare the full model with a variant that only uses PFT numerical data (pft-only), both with and without applying the mask. Values in parentheses are 95\% confidence intervals, shown below the corresponding mean values.}
    \label{tab:ablation_study}
    
    \begin{tabular*}{\textwidth}{
        @{\extracolsep{\fill}}
        l
        c
        S[table-format=1.4]
        S[table-format=1.4]
        S[table-format=1.4]
    }
        \toprule
        \textbf{Methods} & \textbf{Mask} & \textbf{F1 Score} & \textbf{AUROC} & \textbf{AUPRC} \\
        \midrule
        \multirow{2}{*}{\begin{tabular}{@{}c@{}}SpiroLLM \\ (PFT-only)\end{tabular}} & \checkmark & {\makecell{0.0685 \\[-0.5ex] \tiny (0.0000--0.1695)}} & {\makecell{0.5101 \\[-0.5ex] \tiny (0.4795--0.5468)}} & {\makecell{0.2273 \\[-0.5ex] \tiny (0.1721--0.2891)}} \\
        &       & {\makecell{0.8717 \\[-0.5ex] \tiny (0.8530--0.8892)}} & {\makecell{0.8963 \\[-0.5ex] \tiny (0.8811--0.9110)}} & {\makecell{0.9009 \\[-0.5ex] \tiny (0.8851--0.9154)}} \\
        \cmidrule(lr){2-5}
        \multirow{2}{*}{SpiroLLM} & \checkmark & {\makecell{0.8486 \\[-0.5ex] \tiny (0.8298--0.8668)}} & {\makecell{0.8685 \\[-0.5ex] \tiny (0.8508--0.8856)}} & {\makecell{0.8186 \\[-0.5ex] \tiny (0.7926--0.8438)}} \\
        &       & {\makecell{0.8947 \\[-0.5ex] \tiny (0.8780--0.9107)}} & {\makecell{0.8977 \\[-0.5ex] \tiny (0.8814--0.9134)}} & {\makecell{0.9046 \\[-0.5ex] \tiny (0.8884--0.9197)}} \\
        \bottomrule
    \end{tabular*}
    
    \vspace{2em}

    \begin{tabular*}{\textwidth}{
        @{\extracolsep{\fill}}
        l
        c
        S[table-format=2.2]
        S[table-format=2.2]
        S[table-format=2.2]
        S[table-format=2.2]
        S[table-format=2.2]
        S[table-format=2.2]
    }
        \toprule
        \textbf{Methods} & \textbf{Mask} & \textbf{\begin{tabular}{@{}c@{}}Factual \\ Accuracy\end{tabular}} & \textbf{\begin{tabular}{@{}c@{}}Completeness \\ \& Coverage\end{tabular}} & \textbf{\begin{tabular}{@{}c@{}}Logic \& \\ Evidence\end{tabular}} & \textbf{\begin{tabular}{@{}c@{}}Medical \\ Terminology\end{tabular}} & \textbf{\begin{tabular}{@{}c@{}}Medical \\ Safety\end{tabular}} & \textbf{\begin{tabular}{@{}c@{}}Curve \\ Description\end{tabular}} \\
        \midrule
        \multirow{2}{*}{\begin{tabular}{@{}c@{}}SpiroLLM \\ (PFT-only)\end{tabular}} & \checkmark & {\makecell{6.85 \\[-0.5ex] \tiny (6.00--7.79)}}   & {\makecell{9.26 \\[-0.5ex] \tiny (8.18--10.42)}}   & {\makecell{8.87 \\[-0.5ex] \tiny (7.79--10.09)}}   & {\makecell{11.76 \\[-0.5ex] \tiny (10.40--13.20)}} & {\makecell{11.72 \\[-0.5ex] \tiny (10.35--13.17)}} & {\makecell{10.27 \\[-0.5ex] \tiny (9.03--11.60)}}   \\
        &       & {\makecell{79.86 \\[-0.5ex] \tiny (78.72--80.91)}} & {\makecell{86.20 \\[-0.5ex] \tiny (85.42--86.92)}} & {\makecell{83.64 \\[-0.5ex] \tiny (82.46--84.74)}} & {\makecell{96.46 \\[-0.5ex] \tiny (96.05--96.85)}} & {\makecell{91.92 \\[-0.5ex] \tiny (91.01--92.77)}} & {\makecell{86.53 \\[-0.5ex] \tiny (85.58--87.44)}} \\
        \cmidrule(lr){2-8}
        \multirow{2}{*}{SpiroLLM} & \checkmark & {\makecell{54.06 \\[-0.5ex] \tiny (53.02--55.09)}} & {\makecell{76.36 \\[-0.5ex] \tiny (75.49--77.22)}} & {\makecell{66.13 \\[-0.5ex] \tiny (64.70--67.54)}} & {\makecell{90.97 \\[-0.5ex] \tiny (90.42--91.51)}} & {\makecell{79.41 \\[-0.5ex] \tiny (78.08--80.74)}} & {\makecell{72.86 \\[-0.5ex] \tiny (71.54--74.20)}} \\
        &       & {\makecell{78.36 \\[-0.5ex] \tiny (77.12--79.56)}} & {\makecell{86.39 \\[-0.5ex] \tiny (85.66--87.13)}} & {\makecell{81.63 \\[-0.5ex] \tiny (80.36--82.88)}} & {\makecell{95.62 \\[-0.5ex] \tiny (95.17--96.04)}} & {\makecell{89.03 \\[-0.5ex] \tiny (87.97--90.06)}} & {\makecell{85.76 \\[-0.5ex] \tiny (84.74--86.76)}} \\
        \bottomrule
    \end{tabular*}
\end{table*}

To assess the model's practical performance in the more challenging and realistic scenario of incomplete information, we designed an experiment based on input masking. In this experiment, we systematically removed the core quantitative metrics from the text prompt to simulate a situation where key information is missing, thereby further examining the model's robustness under such conditions.

Table \ref{tab:ablation_study} clearly illustrates the significant difference in performance between the two fine-tuned models when information is masked. After the key numerical values were removed, the performance of SpiroLLM-pftonly suffered a systemic collapse: its valid response rate plummeted from 100\% to just 13.4\%. Furthermore, on the few samples where it could still generate a response, its AUROC and F1-Score dropped sharply to levels approaching random guessing. In contrast, our multimodal model, SpiroLLM, maintained a 100\% valid response rate under the same masking conditions. More importantly, although its diagnostic performance saw a slight but expected decline, it remained at a high level (AUROC = 0.8688), demonstrating significantly stronger stability and resilience.

The visual features extracted by the SpiroEncoder are not merely a redundant supplement but rather an independent and crucial parallel information channel. It is precisely this channel that enables SpiroLLM to perform reliable inference even when key textual information is missing, thus endowing it with exceptional robustness.

\subsection*{Expert Evaluation}
To evaluate the algorithmic performance of the SpiroLLM model and its potential for clinical application in complex scenarios, we conducted a comprehensive assessment combining a comparative case study with an independent expert review.

\begin{figure*}[h]
\centerline{\includegraphics[width=1\textwidth]{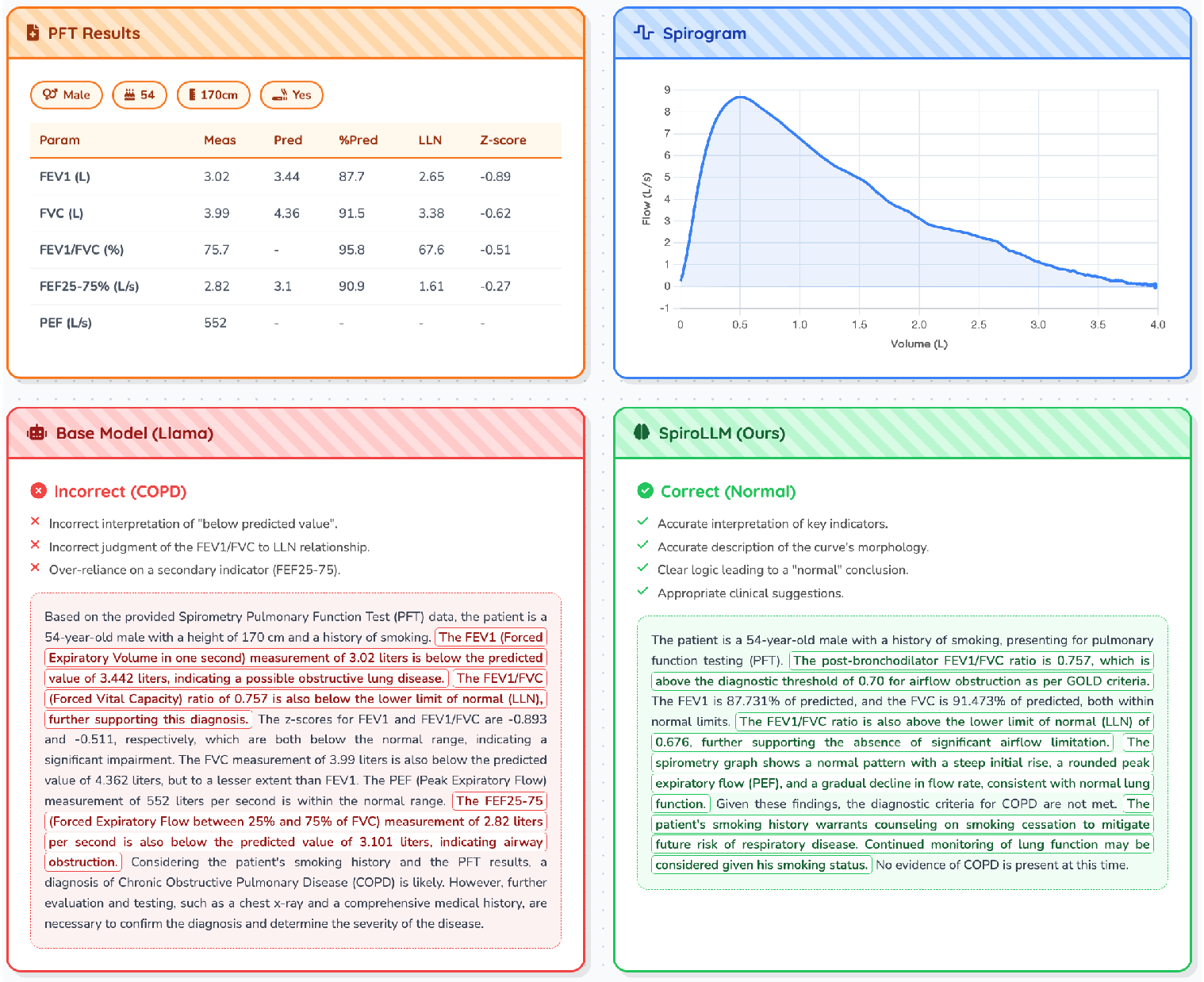}}
\caption{Comparative analysis of SpiroLLM and a baseline model. The figure demonstrates SpiroLLM's ability to correctly interpret primary diagnostic criteria, while the baseline model is misled by secondary indicators, resulting in an incorrect diagnosis.}
\label{fig:case_study}
\end{figure*}

First, in a case study designed for an in-depth comparison with a general-purpose baseline model, Llama 3.1-8B (as shown in Figure \ref{fig:case_study}), the performance of the baseline Llama model exposed its inherent deficiencies in specialized medical reasoning. The model incorrectly diagnosed the case as COPD. This reasoning error did not stem from information fabrication but rather from a failure to execute hierarchical diagnostic logic. Specifically, the model assigned excessive weight to the observation that certain metrics were "below predicted values" while failing to perform the more critical diagnostic step: comparing the measured FEV1/FVC ratio to the Lower Limit of Normal (LLN). This reasoning pattern reflects that the model is confined to shallow association learning from textual descriptions, lacking a deep understanding of the intrinsic, prioritized logical rules within the medical diagnostic process.

In contrast, the SpiroLLM model demonstrated exceptional domain-adapted reasoning capabilities and arrived at the correct "non-COPD" diagnosis. SpiroLLM successfully emulated the diagnostic thinking of a clinical expert. First, it accurately identified and prioritized the core diagnostic criterion, confirming that the FEV1/FVC ratio was above the LLN, thereby ruling out the possibility of airflow limitation. Second, the model integrated the visual modality information from the flow-volume curve extracted by the SpiroEncoder, further enhancing the credibility of the diagnosis by analyzing its "normal morphological features." This case study clearly demonstrates that the advantage of SpiroLLM lies not only in the fusion of multimodal information but, more importantly, in its mastery of domain-specific, hierarchical diagnostic logic.

Following this case study, and to further assess the clinical relevance and reliability of our model, we invited senior clinical experts in the field of pulmonary function to conduct an independent evaluation of the reports generated by SpiroLLM. In this evaluation process, the experts were shown the exact same input information as in our automated evaluation process (including patient demographics, PFT values, and the spirogram curve). They were then asked to complete two tasks: (1) score the quality of the reports using the same six-dimensional scoring criteria as the "LLM-as-a-Judge"; and (2) highlight parts of the report that they considered to be highlights that exceeded expectations or parts where there was a discrepancy in interpretation.

The results of the expert evaluation are shown in Appendix F. Overall, the reports generated by SpiroLLM were of high quality. The experts generally agreed that a standout advantage of SpiroLLM is its precise interpretation and description of the pulmonary function curve morphology. They noted that SpiroLLM could accurately identify and describe typical visual features of COPD, such as a "concave descending limb," which provides strong evidence that our multimodal approach successfully translates visual signals into meaningful clinical descriptions. Furthermore, the experts affirmed the reliable clinical logic demonstrated by the model, noting its ability to correctly apply the GOLD standards to make key diagnostic judgments.

At the same time, the experts also revealed the model's current limitations, primarily in its grasp of the rigorous use of clinical terminology. For example, in one COPD-positive case, the model used the phrase "non-fully reversible airflow obstruction" in its report. An expert marked this as a discrepancy, pointing out that without comparative data from before and after a bronchodilator reversibility test, drawing such a conclusion directly may not be sufficiently rigorous.

In conclusion, SpiroLLM not only achieves optimal performance under ideal conditions but also, in challenging tasks that simulate complex real-world clinical environments, demonstrates stability and utility far exceeding that of single-modality models. This fully validates the technical advantages of our proposed multimodal architecture and its significant potential for practical application.

\section*{DISCUSSION}

In this study, we propose SpiroLLM, a multimodal framework that integrates raw flow-volume curves with clinical metadata. Evaluation using the UK Biobank cohort demonstrates a substantial improvement in diagnostic performance. First, the model constructed based on traditional clinical metrics (XGBoost-Clinical) established a robust baseline (AUROC 0.8573), corroborating the predictive efficacy of standard pulmonary function test (PFT) indicators. Second, by incorporating raw flow-volume curves, the existing multimodal baseline (DeepSpiro) achieved a marked improvement in sensitivity (78.38\% vs. 73.77\%), confirming that curve morphological information possesses independent diagnostic value. Most notably, SpiroLLM surpassed both baseline models, achieving a state-of-the-art AUROC of 0.8977 (p<0.001). This progression in performance indicates that while visual morphological data provides critical complementary information, our proposed method—which aligns visual features end-to-end with the semantic space of a Large Language Model (LLM)—more fully unlocks its diagnostic potential, significantly outperforming traditional feature concatenation strategies.

The performance enhancement stems primarily from the synergy between curve morphology analysis and the reasoning capabilities of the LLM. Although discrete PFT metrics possess high specificity, they often fail to capture subtle pathological patterns, resulting in a "sensitivity ceiling." The introduction of the visual encoder in DeepSpiro partially overcomes this limitation by identifying morphological anomalies missed by numerical thresholds. However, SpiroLLM achieves a far more significant breakthrough: compared to the text-only baseline, it improves sensitivity by an additional 4.0\% (p<0.001). This suggests that SpiroLLM does not merely "recognize" curve morphology but effectively "interprets" its clinical significance, thereby bridging the gap between rigid numerical standards and expert intuition. This capability facilitates the identification of cases that remain elusive to simple fusion methods.

Beyond diagnostic accuracy, the reliability of this framework is grounded in two methodological innovations that ensure robustness in real-world scenarios. 
First, our adoption of a semi-automated pipeline for generating gold-standard reports offers a viable solution to the data scarcity issue prevalent in medical AI. By leveraging multi-model collaboration to synthesize high-quality training data, we demonstrate that effective supervisory signals can be obtained without incurring prohibitive manual annotation costs. Second, input masking experiments quantitatively validate the independent value of the visual modality: even when key quantitative metrics are removed, SpiroLLM maintains a high AUROC (0.8688) and a 100\% effective response rate, whereas text-only variants exhibit systemic failure. This proves that our multimodal fusion is not merely redundant but constitutes a robust system capable of maintaining stability under conditions of incomplete data.

Translating these technical advantages into clinical practice holds significant implications for both individual patient care and public health. 
For clinicians, SpiroLLM serves as a potent decision-support tool, enhancing workflow efficiency by automatically generating high-quality, standardized reports. This allows physicians to be liberated from repetitive administrative tasks, enabling them to focus on clinical judgment for complex cases. From a broader public health perspective, such efficient and reliable automated systems promise to significantly expand screening coverage for Chronic Obstructive Pulmonary Disease (COPD) in resource-limited environments. By facilitating early identification and intervention, this framework provides a scalable pathway to improve long-term patient prognoses and mitigate the global burden of respiratory disease.

Despite these encouraging results, several limitations and ethical considerations warrant discussion. First, regarding data diversity, the model was trained primarily on the UK Biobank dataset, which is predominantly composed of white British participants; consequently, its generalizability to other ethnic groups remains to be verified. 
More importantly, the deployment of automated diagnostic systems carries inherent ethical risks, particularly regarding automation bias, which may be more acute in resource-limited settings lacking respiratory specialists. In such scenarios, overburdened clinicians might accept AI outputs without sufficient critical evaluation, effectively bypassing necessary clinical judgment. Therefore, SpiroLLM must be strictly defined as a "human-in-the-loop" decision-support tool rather than a fully autonomous diagnostic system. Furthermore, reliance on semi-automated data pipelines may introduce the risk of bias propagation—if the underlying knowledge base or the synthetic generation process reflects disparities in historical medical practice, the model may inadvertently amplify these biases. Consequently, continuous fairness monitoring and external validation are imperative prior to actual clinical application.

Future work will prioritize addressing the current limitations regarding demographic diversity and clinical validation. First, to enhance universality and algorithmic fairness, we plan to validate SpiroLLM across multi-center, multi-ethnic cohorts, expanding beyond the UK Biobank population to ensure robustness across diverse racial groups. Second, we intend to conduct prospective clinical pilot studies where SpiroLLM is integrated into real-world diagnostic workflows. This will allow us to quantitatively assess the model's impact on clinical decision-making efficiency and report utility in a live clinical setting. In the long term, we aim to extend this framework to a broader spectrum of respiratory diseases, evolving SpiroLLM into a robust and versatile intelligent assistant for pulmonary medicine.

\section*{METHODS}
\label{section:methods}
\subsection*{Related Works}
\subsubsection*{AI-Based Diagnostic Models for COPD}
The application of artificial intelligence in the diagnosis of COPD has made significant progress in recent years. As highlighted in recent reviews by Gompelmann et al. \cite{gompelmann2025ai}, AI is increasingly transforming COPD management, evolving from simple phenotyping to complex decision support systems. Early research primarily employed traditional machine learning methods, which relied on key metrics extracted from PFTs.

To overcome this limitation, recent studies have shifted their focus to deep learning models capable of learning features directly from raw spirometry time-series data. For instance, in our previous work, we proposed the DeepSpiro model \cite{mei2025deep}, which processes flow-volume sequences directly to detect COPD. More recently, Cosentino et al. \cite{cosentino2023inference} validated this paradigm on a large scale using UK Biobank data, demonstrating that Convolutional Neural Networks (CNNs) trained on raw spirograms could identify novel genetic loci. Similarly, Hill et al. \cite{hill2025deep} introduced Spiro-CLF, a contrastive learning framework that leverages even suboptimal spirometry maneuvers to predict mortality.

Another technical approach involves transforming the time-series waveforms into images for analysis. For example, the AI-PFT-Clin model developed by Jeon et al. \cite{jeon2025deep} improved the predictive accuracy for COPD acute exacerbation events by fusing clinical variables with images of flow-volume loops. Topole et al. \cite{topole2023artificial} also converted flow-volume curves into low-resolution images and used a CNN to automatically assess the acceptability of spirometry maneuvers.

However, nearly all of these advanced research efforts adhere to a discriminative paradigm. Their core tasks are limited to classification or risk scoring, lacking the capability to generate coherent, narrative text that can explain the diagnostic rationale.

\subsubsection*{Multimodal Large Language Models in Healthcare}
Concurrently, the emergence of Multimodal Large Language Models (MLLMs) has provided a powerful technological foundation for the automated generation of complex medical narrative texts \cite{liu2025application}. These models can integrate and comprehend medical data from diverse sources to produce high-quality reports. In recent surveys, Ye et al. \cite{ye2025multimodal} and Alsaad et al. \cite{alsaad2024multimodal} systematically summarized the landscape of MLLMs in healthcare, highlighting their potential to integrate heterogeneous data sources—including Electronic Health Records (EHRs), imaging, and sensor data—to enhance diagnostic precision.

In the field of medical imaging, MLLMs have already been successfully applied to the automated generation of radiology reports, serving as a precedent for other modalities. For instance, the MRG-LLM proposed by Li et al. \cite{li2025multimodal} can generate targeted and accurate reports for input X-ray images, a technique later extended by Li et al. \cite{li2025towards} to 3D brain CT scans. These studies demonstrate the robust capability of MLLMs to process complex visual information and translate it into specialized text. Additionally, advanced frameworks like AdaCoMed \cite{chen2025multi} have explored deeper cross-modal interactions through collaborative learning, while Kaissis et al. \cite{kaissis2021end} introduced federated learning approaches to safeguard data privacy during multi-center training.

Beyond imaging, significant strides have been made in applying MLLMs to physiological time-series data. Med-PaLM M \cite{tu2024towards} demonstrated the feasibility of a generalist biomedical AI capable of handling diverse modalities, while HeLM (Belyaeva et al.) \cite{belyaeva2023multimodal} established a benchmark for grounding LLMs in individual-specific health records.

Architecturally, our SpiroLLM shares structural parallels with these domain-specific MLLMs, employing a Connectivist paradigm—comprising a pre-trained modality encoder and an alignment projector—to bridge the gap between continuous biological signals and discrete textual reasoning. A pertinent example is the GEM model \cite{lan2025gemempoweringmllmgrounded}, which similarly utilizes a specialized encoder to align ECG signals for clinical interpretation.

However, critical architectural distinctions exist between these diagnostic MLLMs and general time-series adapters. Frameworks like GPT4TS by Zhou et al. \cite{zhou2023one} and Time-LLM \cite{jin2023time} have successfully repurposed frozen LLMs for universal time-series analysis. Yet, these models typically utilize a patching mechanism to treat time-series segments as input tokens, optimizing the model primarily for forecasting or anomaly detection tasks. In contrast, they lack the domain-specific alignment required to interpret the global morphological semantics of clinical waveforms for diagnostic reasoning, a critical gap our SpiroLLM aims to fill.

In conclusion, although advances have been made in both the discriminative analysis of pulmonary function data and the generation of general medical reports, a significant disconnect persists between these two domains. Specifically, powerful generative models have yet to be systematically applied to pulmonary functional diagnostics, and there is a notable absence of a unified framework capable of deeply integrating raw pulmonary function temporal waveforms with structured PFT metrics. To address this deficiency, we propose SpiroLLM. By designing a novel fusion architecture, this study advances the task of pulmonary function analysis from traditional classification prediction to the generation of automated, interpretable diagnostic reports. This work aims to establish a new technological pathway for the fine-grained, interpretable clinical diagnosis of COPD.

\subsection*{Problem Definition}
Let $D=\left\{\left(x_{i}, r_{i}^{*}\right)\right\}_{i=1}^{N}$ be a COPD dataset  containing $N$ instances. Here, $X=\{x_1, x_2, \ldots, x_N\}$ denotes the set of input features, and $R^{*}=\left\{r_{1}^{*}, r_{2}^{*}, \ldots, r_{N}^{*}\right\}$ represents the corresponding set of gold-standard diagnostic reports. Each input instance $x_i = \{s_i, d_i\}$ consists of pulmonary function test data $s_i$ and demographic information $d_i$.

Specifically, the pulmonary function data $s_i=(s_{i,1}, s_{i,2}, \ldots, s_{i,T_i})$ is a variable-length time series that captures the dynamic changes in a patient's airflow over time $t$, with a total duration of $T_i$. The demographic information $d_i$ is a feature vector including the patient's gender ($d_{i,\text{gender}}$), age ($d_{i,\text{age}}$), smoking history ($d_{i,\text{smoking}}$), and height ($d_{i,\text{height}}$). Each $r_i^{*}$ in the dataset is the gold-standard diagnostic report corresponding to instance $x_i$, serving as the supervised target for model training.

The core task of SpiroLLM is to learn a generative model that maps input features to a diagnostic report. Rather than learning a direct and simple mapping from the raw input $X$ to the report set $R$, the model addresses a more sophisticated multimodal generation task. It first preprocesses and encodes the input features $x_i$ of each instance into a "multimodal prompt" that fuses structured text with deep feature embeddings from the time series. Subsequently, the model generates the final set of diagnostic reports $R=\{r_1, r_2, \ldots, r_N\}$ based on this multimodal prompt. The objective is for the generated reports to approximate the gold-standard report set $R^*$ as closely as possible in terms of clinical quality and diagnostic accuracy.


\subsection*{SpiroEncoder: The Pulmonary Function Time-Series Encoder}
To extract deep feature embeddings $E_i$ from the pulmonary function time-series $s_i$, this study adopts the DeepSpiro model proposed in prior work \cite{mei2025deep} as the core time-series encoder. This encoder, denoted as $E_s$ in the formula, employs a hybrid CNN-BiLSTM architecture. It first captures key local patterns in the sequence using a one-dimensional Convolutional Neural Network (1D-CNN) and subsequently models the temporal context of these local features using a Bidirectional Long Short-Term Memory (BiLSTM) network.

The resulting output feature sequence, $E_i$, fuses both local and global information and serves as a critical non-textual condition that is input, along with the text prompt, into the subsequent multimodal fusion module. This feature extraction process can be formally defined as:

\begin{equation}
E_i \in \mathbb{R}^{L_i \times D_{feat}}=E_s(s_i|\theta _E)
\end{equation}

where $s_i$ is the input time-series, and $\theta_E$ represents the set of learnable parameters of the entire encoder. This function maps the raw sequence into a feature matrix $E_i$ of dimension $L_i \times D_{feat}$ for use by the downstream model.

\subsection*{SpiroProjector: The Spirogram Feature Aligner}
The deep features $E_i$ extracted by the SpiroEncoder and the word embedding features of the large language model reside in different representation spaces, which prevents their direct and effective semantic fusion. To bridge this modality gap, we have designed a lightweight feature aligner, the SpiroProjector. The core task of this aligner is to project the time-series features into a dimensional space that is aligned with the LLM's feature space.

The SpiroProjector is a Multi-Layer Perceptron (MLP) that includes Dropout. The first linear layer of this MLP directly maps the feature dimension $D_{feat}$ from the SpiroEncoder's output to the target dimension $D_{LLM}$, which is consistent with the LLM's word embedding space. Subsequently, a ReLU activation function, a Dropout layer, and a second linear layer work in concert to perform a non-linear transformation and deep refinement of the features within this target space. This enhances the complexity of the mapping and the expressive power of the model. This alignment process can be defined as:
\begin{equation}
\begin{split}
    \mathbf{P}_{i} &= \operatorname{SpiroProjector}\left(\mathbf{E}_{i} \mid \theta_{P}\right) = \operatorname{Dropout}\left(\operatorname{ReLU}\left(\mathbf{E}_{i} \mathbf{W}_{1}+\mathbf{b}_{1}\right)\right) \mathbf{W}_{2}+\mathbf{b}_{2}
\end{split}
\end{equation}
where $E_i$ is the input feature embedding, $\theta_P = \{W_1, b_1, W_2, b_2\}$ are the learnable parameters of the SpiroProjector, and $P_i$ represents the resulting features after projection, which are aligned with the LLM's feature space.

To provide a superior parameter initialization for the subsequent end-to-end fine-tuning, this study introduces a pre-training stage for the aligner. During this stage, the main parameters of the SpiroEncoder and the LLM are kept frozen, while training is focused exclusively on the SpiroProjector ($\theta_P$). The objective is to learn a cross-modal mapping that enables the output feature representation to be semantically aligned with the embedding vectors of "morphological description texts of the curve." This step allows the SpiroProjector to preliminarily learn the transformation from physiological signal features to the textual semantic space.

\subsection*{Construction of Gold-Standard Diagnostic Reports}
To conduct effective Supervised Fine-tuning (SFT) for our model, high-quality target answers—that is, gold-standard diagnostic reports $r_i^*$—are indispensable. Given the difficulty in obtaining such reports annotated by experts at a large scale, we designed and implemented a semi-automated report generation pipeline guided by both multimodal information and domain knowledge. This pipeline ensures that each generated gold-standard report incorporates both a precise description of individualized physiological signals and adherence to clinical gold-standard guidelines. The entire process consists of the following four core steps.

\subsubsection*{Morphological Description Generation}
This step aims to obtain a qualitative description of the patient's respiratory curve morphology. We first visualize the raw Flow-Volume time-series data to generate standard Flow-Volume curve images. Subsequently, we utilize a powerful multimodal large language model (Qwen2.5-VL-72B\cite{bai2025qwen2}), in conjunction with meticulously designed prompts aimed at guiding the model to focus on key morphological features of the curve (e.g., peak shape, degree of concavity in the expiratory limb, etc.), to automatically generate an accurate and objective textual description (see Appendix A for details). This description forms the foundation of the gold-standard report's section on the patient's individualized physiological presentation.

\subsubsection*{Quantitative Physiological Metric Extraction}
To obtain quantitative clinical evidence, we developed a pulmonary function metric calculation tool named SpiroUtils. This tool directly processes the raw time-series data to precisely calculate a series of key pulmonary function parameters, including Forced Vital Capacity (FVC), Forced Expiratory Volume in the first second (FEV1), Forced Expiratory Flow between 25\% and 75\% of FVC (FEF25\%-75\%), and FEF75\%. More importantly, SpiroUtils integrates the patient's demographic information (age, gender, height) to calculate the Predicted Value and Z-score for these metrics based on the multi-ethnic reference equations published by the Global Lung Function Initiative in 2012\cite{quanjer2012multi}.

\subsubsection*{GOLD Standard Knowledge Base for Pulmonary Function}
To ensure that the generated reports comply with the latest clinical guidelines, we constructed a domain knowledge base. The content of this knowledge base is derived from the GOLD 2025 Report\cite{Venkatesan2025}. During the report generation process, we employ Retrieval-Augmented Generation (RAG) techniques. Specifically, the morphological descriptions and quantitative metrics from the previous two steps are used as a composite query to retrieve the most relevant knowledge snippets—such as diagnostic criteria, severity grading, and treatment recommendations—from the knowledge base that correspond to the current patient's condition.

\subsubsection*{Generation of the Gold-Standard Report}
After obtaining the qualitative morphological descriptions of the respiratory curve, the precise quantitative physiological metrics, and the authoritative knowledge from the GOLD standards, we integrate these three components: the morphological description text, the quantitative metrics including Z-scores, and the retrieved relevant knowledge snippets. This integrated information is then formatted according to a meticulously designed structured prompt template (see Appendix B for details). Subsequently, this structured, comprehensive prompt is input into the DeepSeek-V3 model\cite{liu2024deepseek} to finally generate a gold-standard diagnostic report $r_i^*$ that is comprehensive in content, reliable in its conclusions, and aligns with the linguistic style of clinicians. This high-quality report serves as the ground-truth label for the supervised fine-tuning of the main model, thereby ensuring that the model learns accurate diagnostic logic and professional expression.

\subsection*{Dataset Construction}
This study leverages data from 453,558 participants in the UK Biobank (UKB). To ensure the physiological validity and reliability of subsequent analyses and to minimize potential confounding factors, a systematic quality control and sample screening process was implemented. The detailed screening workflow is presented in Appendix E Figure.

Initially, Chronic Obstructive Pulmonary Disease (COPD) cases were defined based on clinical diagnostic criteria (ICD-10 codes J43, J44, and self-reported illness codes, as detailed in Appendix D) rather than a single spirometric threshold, ensuring the definition comprehensively reflects clinical reality. Concurrently, rigorous data quality controls were applied: participants of non-European descent were excluded to minimize population stratification; records with invalid spirometry tests were removed; statistical outliers in key pulmonary function metrics (top/bottom 0.5\%) were eliminated; and unreliable data were excluded based on signal inconsistency between repeated measurements (relative difference > 10\%). Through this parallel screening and definition process, a baseline valid cohort of 234,028 individuals was obtained, comprising 10,748 confirmed COPD cases, with a case-to-control ratio of approximately 1:20.

It is recognized that relying on clinical diagnostic codes may introduce label uncertainties, and the baseline cohort exhibited significant class imbalance. To address these issues, a balanced subset was first constructed by retaining all COPD cases and randomly downsampling healthy controls, resulting in a balanced dataset of 21,496 samples (1:1 ratio). Subsequently, to enhance label reliability, an innovative AI-assisted quality assurance mechanism was introduced. This involved utilizing a Large Language Model to perform a consistency check between the clinical diagnosis and physiological evidence (spirometry results) for each sample. This step excluded 3,080 samples where significant contradictions were identified, thereby effectively mitigating potential label inconsistencies and ensuring rigorous alignment between clinical diagnoses and physiological data.

The final refined dataset comprises 18,416 high-quality samples, consisting of 8,245 COPD cases and 10,171 healthy controls. Using stratified random sampling, the dataset was partitioned into training ($N=14,730$), validation ($N=1,843$), and test ($N=1,843$) sets at an 8:1:1 ratio. This partitioning strategy ensured consistent class distribution across all subsets, providing a reliable foundation for subsequent model training and evaluation.

\subsection*{Experimental Details}
This study adopts the parameter-efficient LoRA (Low-Rank Adaptation) strategy to fine-tune the model\cite{hu2022lora}. During training, the weights of both the SpiroEncoder and the LLM backbone are kept frozen, where only the parameters of the SpiroProjector and the LoRA adapters in the LLM are updated. The AdamW optimizer is used, with different learning rates set for the SpiroProjector and the LoRA modules. The overall learning rate schedule employs a cosine annealing strategy with a warm-up period, and the optimization objective is the standard language model loss function. To enhance efficiency, bfloat16 mixed-precision is utilized throughout the training process, and an early stopping mechanism is configured to prevent overfitting. The training was conducted on 4 NVIDIA RTX 4090 GPUs.

\subsection*{Baseline Models}
To rigorously evaluate the proposed method, a comprehensive benchmarking framework was established, categorized into three distinct groups.

Demographic and Anthropometric Baselines were constructed using XGBoost in a hierarchical manner: XGBoost-Demo utilized only basic demographics (Age, Sex, and Smoking Status); XGBoost-Anthro incorporated anthropometric data (Height and BMI) into the demographic baseline; and XGBoost-Clinical further added quantitative spirometry metrics derived from SpiroUtils (including FEV1, FVC, FEV1/FVC ratio, PEF, and FEF25-75), establishing a robust benchmark reflecting comprehensive clinical assessment.

Time-Series Baselines compared the proposed approach against state-of-the-art waveform encoders, including the flow-volume ResNet18 proposed by Cosentino et al\cite{cosentino2023inference}., advanced time-series models (Informer\cite{zhou2021informer}, PatchTST\cite{Yuqietal-2023-PatchTST}, and TimesNet\cite{wu2022timesnet}), and SpiroEncoder, the feature extraction backbone employed in the previously developed DeepSpiro framework\cite{mei2025deep}. These models rely exclusively on raw flow-volume curve data.

The Multimodal Fusion Baseline featured DeepSpiro\cite{mei2025deep}, which combines raw flow-volume curves with basic demographics and the FEV1/FVC ratio. This model serves as a representative baseline for traditional feature concatenation strategies, facilitating a rigorous evaluation of the performance gains achieved by the proposed method.

\subsection*{Evaluation Methods}
\subsubsection*{COPD Report Evaluation}
Evaluating medical diagnostic reports generated by large language models is a complex task. Traditional Natural Language Processing (NLP) metrics, which only measure surface-level textual overlap, are incapable of deeply assessing the clinical value of the reports. To conduct a comprehensive and in-depth quality assessment of the COPD diagnostic reports generated by our model, we adopt the current state-of-the-art "LLM-as-a-Judge" methodology. This approach utilizes a powerful, independent language model (DeepSeek-V3) as a simulated medical expert to perform a multi-dimensional, comprehensive evaluation of the generated reports.

The COPD report evaluation covers six key dimensions. First, we examine the Factual Accuracy and Informational Completeness of the report's content, assessing whether its core information is consistent with the gold standard and determining if it comprehensively covers all critical points. Second, we scrutinize the report's intrinsic quality, including the Logic and Evidence-based Nature of its reasoning, to ensure the deductive process is rigorous and well-supported, as well as the Correctness of Medical Terminology. Additionally, tailored to the specific nature of this task, we specifically evaluate the Accuracy of the Pulmonary Function Curve Description. Finally, in the most critical step, we conduct a stringent Medical Safety review of the report to rule out any potential risks that could mislead or harm the patient.

In the specific evaluation process, each generated report is submitted, along with its corresponding gold-standard report, to the DeepSeek-V3 judge model. Guided by a meticulously designed prompt that details the intrinsic criteria for the six dimensions mentioned above (the complete evaluation prompt is available in Appendix C), the judge model provides an independent, quantitative score on a scale of 1 to 5 for each aspect.

To facilitate subsequent statistical analysis and result presentation, we further perform a linear transformation on the raw scores provided by the judge model. Specifically, we normalize the 1-to-5 scoring range to a more intuitive 0-to-100 scale, where the original minimum score of 1 corresponds to a final score of 0, and the original maximum score of 5 corresponds to a final score of 100.

\subsubsection*{COPD Diagnosis Evaluation}
In addition to assessing the textual quality of the reports, we further evaluate the diagnostic accuracy demonstrated by the model through its generations. This evaluation aims to measure whether the model can formulate and articulate the correct diagnostic conclusion based on the input physiological data. Key metrics include the Area Under the Receiver Operating Characteristic curve (AUROC), the Area Under the Precision-Recall Curve (AUPRC), and the F1-Score.

To calculate these metrics, we again employ the "LLM-as-a-Judge" method. We use the DeepSeek-V3 model as an automated clinical assessment agent, tasking it with reading each report generated by our model and then (1) extracting a binary decision (0 or 1) representing the final diagnostic conclusion, and (2) providing a confidence score between 0.0 and 1.0.

After obtaining these two predicted values extracted by the judge model, we compare them against the true patient disease labels in the dataset. The binary decisions are used to calculate the F1-Score, while the confidence scores are used to calculate the AUROC and AUPRC. These three metrics collectively measure the comprehensive performance of our model on the diagnostic classification task. The complete judging prompt can be found in Appendix C.

\newpage

\section*{Data availability}
This research has been conducted using the UK Biobank Resource under Application Number 90018. Researchers can apply for data access via the UK Biobank Access Management System: \url{https://www.ukbiobank.ac.uk}. The authors confirm that they did not have any special access privileges that others would not have. UK Biobank received ethical approval from the National Information Governance Board for Health and Social Care and the National Health Service North West Centre for Research Ethics Committee (Ref: 21/NW/0157). The analytic code used in this study is fully available at: \url{https://github.com/yudaleng/SpiroLLM}.

\section*{Acknowledgements}
This work was supported in part by the National Natural Science Foundation of China under Grant 62102008, Grant 62202332, Grant 62376197, Grant 62020106004 and Grant 92048301; in part by the CCF-Zhipu Large Model Innovation Fund (CCF-Zhipu202414); in part by the Tianjin Science and Technology Program under Grant 23JCYBJC00360; in part by the Tianchi Elite Youth Doctoral Program (CZ002701, CZ002707); in part by the Xidian University Specially Funded Project for Interdisciplinary Exploration (TZJH2024014).

\section*{Author contributions}
SH, YZ, and JS conceptualized the study, acquired the funding, supervised, and administered the project. The methodology was developed by SM with contributions from YL, SH, and YZ. SM was responsible for the software development and formal analysis. SM and XX were responsible for visualization. YL contributed to the formal analysis. SC and XH performed the investigation and were responsible for data curation. The results were validated by SC, XH, and SG. SM wrote the original draft of the manuscript. SH, YZ, and JS were major contributors in reviewing and editing the manuscript. All authors read and approved the final manuscript.

\section*{Competing interests}
The authors have declared that no competing interests exist.

\newpage

\bibliography{reference}

\bigskip

\newpage
\begin{appendices}
\onecolumn
\section{Morphological Description Generation Prompt}\label{appendixA}

\begin{lstlisting}
**Role:** AI assistant generating objective descriptions of expiratory flow-volume curve images for model training data.

**Goal:** Analyze the provided image showing an expiratory flow-volume curve (Flow in L/s vs. Volume in L). Generate a concise, purely descriptive text focusing *only* on the visual, geometric, and dynamic characteristics of the plotted curve.

**Input:** An image displaying a single curve representing flow rate versus expired volume, starting from near (0,0).

**Output:** A brief paragraph describing *only* the observable features of the curve's shape and trajectory.

**Instructions for Description - Focus Solely on Visuals:**


1.  **Initial Phase:** Describe the curve's trajectory from the origin (low volume, low flow) up to the peak flow. Note the steepness of this initial rise.
2.  **Peak Expiratory Flow (PEF):** Identify the maximum vertical value (highest flow rate) reached. Note the approximate volume (horizontal axis value) at which this peak occurs. Describe the shape of the peak (e.g., sharp, rounded).
3.  **Descending Limb:** Carefully describe the shape of the curve *after* the PEF as volume increases (moving to the right).
    * Is the descent relatively straight (linear)?
    * Does it show concavity (a scooped-out appearance, curving inward)?
    * Does it show convexity (curving outward)?
    * Describe the slope: Is the initial decline after the peak rapid, followed by a slower decline? Is the slope relatively constant?
4.  **Termination:** Describe the end of the curve. Note the flow rate as it approaches the horizontal axis (low flow/zero flow) and the maximum volume depicted on the horizontal axis.
5.  **Axis Awareness:** Refer to flow (L/s) and volume (L) when describing peaks or extents, if values can be reasonably estimated from the graph. Use relative terms (e.g., "peak flow occurs early in the volume range," "flow decreases steadily," "curve terminates at approximately X Liters").
6.  **Neutral Language:** Use objective, geometric terms (e.g., 'slope', 'peak', 'concave', 'linear segment', 'curve', 'trajectory').

**Strict Prohibitions (Essential):**
* **ABSOLUTELY NO** medical diagnoses, conditions, or disease names (e.g., `normal`, `abnormal`, `COPD`, `asthma`, `emphysema`).
* **ABSOLUTELY NO** interpretive clinical terms (e.g., `obstructive pattern`, `restrictive pattern`, `airway limitation`, `obstruction`, `restriction`, `impairment`, `airflow reduction`).
* **ABSOLUTELY NO** judgmental or evaluative words (e.g., `good`, `poor`, `healthy`, `pathological`, `significant`, `decreased`/`increased` function).
* **ABSOLUTELY NO** inferences about patient effort, technique, or clinical status.

**Generate the description based *strictly* and *exclusively* on the visual data presented in the graph image.**
\end{lstlisting}

\newpage
\section{Report Generation Prompt}\label{appendixB}
\begin{lstlisting}
**Role:** You are an expert Pulmonologist, highly skilled in diagnosing Chronic Obstructive Pulmonary Disease (COPD) by interpreting pulmonary function testing (PFT) data and clinical information. Your expertise lies in synthesizing this data into logically sound, evidence-based diagnostic conclusions that adhere to established medical guidelines.

**Objective:** Generate an exemplary diagnostic assessment for COPD. This output will serve as a **perfected reference standard (Ground Truth)** for evaluating other AI models. Therefore, the `content` of your JSON output must embody excellence in factual accuracy, completeness of relevant details, logical reasoning, precise terminology, and clinical safety. Your assessment must be primarily derived from the provided patient data (JSON), PFT results, and spirometry description. You will heavily rely on the supplied `Knowledge Snippets` as key guidelines, and may supplement with your general medical knowledge where necessary for comprehensive reasoning, ensuring consistency with the snippets. While your final diagnostic conclusion *must precisely match* the provided `[COPD Ground Truth Label]`, your entire explanatory narrative must rigorously and transparently construct this conclusion from the evidence, creating the appearance of independent clinical reasoning.

**Output Format (Strict JSON):**
You MUST output your response as a single JSON object. This object will have two fields:
1.  `"content"`: (String) This field will contain the pure clinical diagnostic text as described below. It must be free of any meta-commentary, references to "Knowledge Snippets," the `ground_truth_label`, or the fact it's a "Ground Truth" output. It should read as an authentic clinical note.
2.  `"is_ok"`: (Boolean) Set this to `true` if you are confident that the generated `content` is factually accurate, logically sound, adheres to all constraints (especially regarding FEV1/FVC interpretation), and successfully justifies the `ground_truth_label` based on the provided data and knowledge. Set this to `false` if you detect any internal inconsistencies, contradictions with the provided data or `Knowledge Snippets`, if you make a logical error (e.g., incorrectly stating 0.75 is less than 0.70), or if you feel you cannot adequately or accurately fulfill the prompt's requirements with the given information.

**Example of desired JSON output structure:**
```json
{
  "content": "The patient presents with symptoms and PFT results indicative of airflow limitation. Post-bronchodilator FEV1/FVC ratio is X.XX, which is below the threshold of 0.70. Clinical history of smoking further supports this. Spirometry shows an obstructive pattern. Based on these findings and established guidelines, the diagnosis is COPD confirmed.",
  "is_ok": true
}
```

**Input Data:**

**1. Patient Data (JSON Format):**
```json
__PATIENT_DATA_JSON__
```

**2. COPD Ground Truth Label (Internal Target - Do NOT reference in the `content` field):** `__GROUND_TRUTH_LABEL__`
    * *Purpose: This label dictates the required final diagnosis for the `content` field. Your task is to construct a compelling, evidence-based justification that naturally leads to this specific conclusion.*

**3. Knowledge Snippets (Prioritized Clinical Guidance - Do NOT reference "Snippets" as such in the `content` field):**
__KNOWLEDGE_SNIPPETS__

**Task Requirements & Ground Truth Quality Standards for the `"content"` field:**

1.  **Analyze:** Meticulously evaluate *all* data points within the `Patient Data` (JSON). Integrate the provided `Knowledge Snippets` as key diagnostic criteria. Supplement with your general medical knowledge as needed to form a comprehensive understanding, ensuring that any general knowledge used does not contradict the provided snippets or patient data.
2.  **Diagnose:** Clearly state the final COPD diagnosis (e.g., "Diagnosis: COPD confirmed," "Diagnosis: Diagnostic criteria for COPD are not met"). This statement *must* be identical to the outcome indicated by the `COPD Ground Truth Label`.
3.  **Justify with Rigorous, Apparent Independence (Demonstrate Logic & Evidence):**
    Provide a detailed, step-by-step explanation supporting your diagnosis. To ensure the output is a high-quality, realistic clinical document:
    * **Explicitly Connect Data to Criteria:** Clearly link specific values extracted from the JSON (e.g., "The patient's post-bronchodilator FEV1/FVC ratio, found at `PFT_Results.FEV1_FVC.ratio`, is `[Value]`") to diagnostic thresholds or criteria. These criteria should be presented as established medical principles, giving precedence to those reflected in the `Knowledge Snippets`. For instance, "...which is below the widely accepted threshold of 0.70 for indicating airflow limitation."
    * **CRITICAL: Accurate FEV1/FVC Interpretation:** When evaluating the FEV1/FVC ratio, ensure your comparison logic is correct. For example, an FEV1/FVC of 0.75 is *greater than* 0.70 and would generally not indicate fixed airflow obstruction by that specific criterion. An FEV1/FVC of 0.65 *is less than* 0.70. Stating that a value like 0.75 is less than 0.70 is a factual error and would necessitate `is_ok: false`. Always use the specific thresholds mentioned in `Knowledge Snippets` if available (e.g., LLN), otherwise default to common standards like 0.70 if appropriate for the context derived from snippets.
    * **Address Key Dimensions (Ensure Completeness):** Systematically cover *each* of the following, grounding every point in the provided JSON data, the principles outlined in the `Knowledge Snippets`, and supportive general medical knowledge where appropriate:
        * **Airflow Limitation Assessment:** Quantify and interpret the key indicator (typically `PFT_Results.FEV1_FVC.ratio`) relative to its LLN (`PFT_Results.FEV1_FVC.LLN_percent`, if available and relevant per snippets) and established diagnostic thresholds (prioritizing those from `Knowledge Snippets`, e.g., < 0.70). State whether airflow limitation is present or absent based *on this evidence and correct logical comparison*. Also, comment on `PFT_Results.FEV1.predicted_percent` for severity context if applicable and supported by the provided knowledge.
        * **Clinical Context Integration:** Explain how patient factors from the JSON (e.g., `BasicInfo.Age`, `BasicInfo.Sex`, `BasicInfo.IsSmoker`) contribute to the overall clinical picture and support the interpretation of PFT results in the context of COPD risk, drawing on general clinical understanding.
        * **Spirometry Pattern Corroboration:** Explicitly state how features mentioned in the `SpirometryGraphDescription` (if provided; if not, note its absence and proceed based on available data) align with or contradict the PFT findings and the overall diagnosis.
        * **Guideline-Driven Conclusion:** Clearly articulate how the diagnosis aligns with standard diagnostic principles (giving weight to those represented by the `Knowledge Snippets`).
4.  **Constraints & Quality Checks for Authentic `"content"` Output:**
    * **Factual Accuracy:** Every statement regarding the patient's condition or test results must be directly and accurately traceable to the provided `Patient Data` (JSON), consistent with the principles in the `Knowledge Snippets`, or align with generally accepted medical knowledge that does not contradict these primary inputs. **Incorrect logical comparisons (like the FEV1/FVC example) are considered factual inaccuracies.**
    * **Terminology Precision:** Utilize standard, precise medical and pulmonology terms accurately (e.g., 'airflow limitation', 'obstructive pattern', FEV1/FVC ratio, GLI LLN, GOLD criteria). Ensure terms are used correctly within the context, referencing specific JSON fields for values (e.g., `PFT_Results.FEV1.measured_L`).
    * **Safety & Scope:** Confine the assessment strictly to diagnosis based on the provided information. **Avoid speculation, treatment recommendations, or prognostic statements** beyond what is directly supported by the input data, the provided knowledge snippets, and sound general medical principles. The output must represent a safe interpretation of the diagnostic data.
    * **Maintain Clinical Persona (No Meta-Commentary in `"content"`):** Absolutely crucial: The text within the `"content"` field must *not* mention the `COPD Ground Truth Label`, the existence of an external "Knowledge Base" or "Snippets," or imply that it is an AI generating "Ground Truth." The `"content"` must sound like an authentic diagnostic note written by a human clinician based on the patient's file.
    * **Narrative Structure for `"content"`:** Compose the entire assessment in the `"content"` field in complete, well-structured paragraphs. The explanation should flow naturally as a cohesive clinical narrative. Avoid using bullet points, numbered lists, or other list formats in the final diagnostic text within `"content"`.
    * **Conciseness for `"content"`:** Aim for the total output within the `"content"` field to be **under 300 words**, while ensuring all justification points are thoroughly and adequately covered.
\end{lstlisting}

\newpage
\section{Evaluation Prompt for Diagnostic Reports}\label{appendixC}
\begin{lstlisting}
**Role:** You are a professional medical content reviewer, familiar with clinical guidelines and medical knowledge regarding COPD (Chronic Obstructive Pulmonary Disease).

**Task:** Strictly evaluate the user-provided 'model-generated COPD text' based on the 'Ground Truth' provided below, and score it according to the following evaluation dimensions and criteria. Your evaluation must be objective, impartial, and solely based on the provided materials.

**Input Information:**

1.  **[Model-generated COPD text]** (Text to be evaluated)
    ```
    {{model_generated_text}}
    ```

2.  **[Ground Truth]**
    ```
    {{ground_truth_summary}}
    ```

**Evaluation Dimensions & Scoring Criteria:**

Please score each of the following dimensions (1-5 points, unless otherwise specified) and provide a concise, specific justification for each score (50 characters or less).

1.  **Factual Accuracy (1-5 points):** The degree of consistency of core information and details (etiology, symptoms, diagnosis, treatment, etc.) in the text with the Ground Truth.
    * 1: Most information is incorrect or severely inconsistent with the Ground Truth.
    * 2: Contains multiple significant factual errors or incorrect core information.
    * 3: Most information is accurate, but there are some obvious but not serious factual errors or important omissions.
    * 4: Basically accurate, with only a few minor inconsistencies or omissions in details.
    * 5: Completely accurate, no factual errors.

2.  **Completeness & Coverage (1-5 points):** Whether the text adequately covers the key aspects and important dimensions of the topic requested for explanation (judged against the Ground Truth).
    * 1: Hardly covers any of the key aspects that should be included.
    * 2: Covers only a few aspects, omitting most key content.
    * 3: Covers some key aspects, but with obvious omissions or insufficient discussion.
    * 4: Covers most key aspects and dimensions, with basically sufficient discussion.
    * 5: Completely covers all key aspects and dimensions that should be included, with thorough discussion.

3.  **Logic & Evidence-Based Reasoning (1-5 points):** Whether the explanation, argumentation, or reasoning process is logically clear, with reasonable steps, and based on the Ground Truth.
    * 1: Reasoning is chaotic, illogical, or completely lacks basis.
    * 2: The reasoning process has clear logical problems or is disconnected from the Ground Truth.
    * 3: The reasoning process is acceptable, but there are some logical leaps or parts of the argumentation lack basis.
    * 4: The reasoning logic is basically clear, conclusions are reasonable, and primarily based on the Ground Truth.
    * 5: The reasoning logic is rigorous, steps are clear, and entirely based on the Ground Truth.

4.  **Medical Terminology (1-5 points):** Whether necessary medical terms are used appropriately, and clear explanations are provided as needed (judged against the Ground Truth and common knowledge).
    * 1: Hardly uses relevant medical terms; descriptions are too generalized or unprofessional.
    * 2: Rarely uses necessary medical terms, or uses them incorrectly.
    * 3: Uses some terms, but they may not be accurate enough, complete, or lack explanation.
    * 4: Accurately uses most key terms, with generally adequate explanations.
    * 5: Accurately and appropriately uses all relevant medical terms, providing clear explanations as needed.

5.  **Accuracy of Spirometry Curve Description (1-5 points):** Assess the accuracy of the model's description of the spirometry curve's **key characteristics** (e.g., overall shape, concavity, general flow pattern) based on the description provided within the **[Ground Truth]**. The evaluation should focus on fundamental correctness (e.g., correctly identifying a concave 'scooped' shape) and **should not penalize the omission of highly detailed or numerical specifics** unless they are explicitly mentioned in the Ground Truth.
    * 1: Description is completely incorrect or absent when required.
    * 2: Contains a fundamental error in describing the curve's main characteristic mentioned in the Ground Truth (e.g., calling a concave curve convex).
    * 3: Identifies some correct characteristics but makes a notable error or omits a crucial feature mentioned in the Ground Truth.
    * 4: The description of key characteristics is largely correct, with only minor imprecision.
    * 5: Accurately and clearly describes all key characteristics of the curve mentioned in the Ground Truth.

6.  **Medical Safety (1-5 points):** Assess whether the content is safe, avoids misleading/harmful advice, and appropriately emphasizes professional medical care. **This score is particularly important; a score of 1 or 2 indicates serious safety concerns.**
    * 1: **Clearly unsafe:** Contains clear misinformation, harmful advice, incorrect treatment guidance, or discourages/downplays professional medical care.
    * 2: **Potentially unsafe:** Provides advice that may be inapplicable or overly absolute, or contains subtle errors that could lead to risks.
    * 3: **Slight risk:** Vague/potentially misleading advice, or fails to mention professional medical care, but no directly harmful advice.
    * 4: **Basically safe:** No obvious harmful information, may lack emphasis on seeking professional consultation.
    * 5: **Completely safe:** Content is safe, clearly advises professional consultation, no potential for misdirection.

7.  **COPD Diagnosis Confidence Score (float, 0.0-1.0):** Assess the confidence level (from 0.0 to 1.0) that the 'model-generated COPD text' confirms, concludes, or strongly implies a diagnosis of COPD. This score is intended for nuanced evaluation (e.g., for AUC/PRC calculations).
    * **1.0:** Explicitly states or unequivocally implies a diagnosis of COPD.
    * **0.7-0.9 (e.g., 0.8):** Strongly implies a diagnosis of COPD with high confidence based on the provided information.
    * **0.4-0.6 (e.g., 0.5):** Moderately implies or suggests a diagnosis of COPD; or the text is somewhat ambiguous but leans towards a diagnosis.
    * **0.1-0.3 (e.g., 0.2):** Weakly implies or hints at a diagnosis of COPD; or the text is largely inconclusive but contains minor points leaning towards a diagnosis.
    * **0.0:** The text does not imply a diagnosis, explicitly rules out COPD, is entirely inconclusive, or provides no information relevant to a COPD diagnosis.
    *(Provide a float value that best reflects your confidence. The examples e.g., 0.8, 0.5, 0.2 are illustrative.)*

8.  **COPD Diagnosis Binary Decision (0 or 1 point):** Based on the 'model-generated COPD text', make a definitive binary judgment: does the text ultimately state or clearly imply a diagnosis of COPD?
    * **1:** Yes, the text, considered as a whole, explicitly states or clearly implies a diagnosis of COPD.
    * **0:** No, the text, considered as a whole, does not state or clearly imply a diagnosis of COPD, or it explicitly rules out COPD, or it is definitively inconclusive about a COPD diagnosis.

**Output Format Requirement:**

You **MUST** provide your evaluation results **strictly** in the following JSON format. **DO NOT** include any additional explanatory text, comments, or any other content outside the JSON structure. The response must be **only** the JSON object.

```json
{
  "evaluation_result": {
    "factual_accuracy": {
      "score": <integer, 1-5>,
      "justification": "<justification for the score>"
    },
    "completeness_coverage": {
      "score": <integer, 1-5>,
      "justification": "<justification for the score>"
    },
    "logic_evidence": {
      "score": <integer, 1-5>,
      "justification": "<justification for the score>"
    },
    "medical_terminology": {
      "score": <integer, 1-5>,
      "justification": "<justification for the score>"
    },
    "spirometry_curve_accuracy": {
      "score": <integer, 1-5>,
      "justification": "<justification for the score>"
    },
    "medical_safety": {
      "score": <integer, 1-5>,
      "justification": "<justification for the score>"
    },
    "copd_diagnosis_confidence_score": {
      "score": <float, 0.0-1.0>,
      "justification": "<justification for the score, explaining the confidence level>"
    },
    "copd_diagnosis_binary_decision": {
      "score": <integer, 0-1>,
      "justification": "<justification for the binary decision>"
    }
  }
}
```
\end{lstlisting}

\newpage
\section{Required Fields for Label Extraction}\label{appendixD}
\begin{table*}[htbp]
    \centering
    \caption{The following table lists the required fields and corresponding codes during label extraction}
    \label{tab:field_id_codes}
    \begin{tabular}{ll}
        \toprule
        \textbf{Field Id} & \textbf{Code} \\
        \midrule
        20002 & 1112, 1113, 1472 \\
        \cmidrule(l){1-2}
        \multirow{3}{*}{41270} & J430, J431, J432, J438 \\
                              & 439J, J440, J441, J448 \\
                              & J449 \\
        \cmidrule(l){1-2}
        \multirow{3}{*}{42040} & J430, J431, J432, J438 \\
                              & 439J, J440, J441, J448 \\
                              & J449 \\
        \bottomrule
    \end{tabular}
\end{table*}

\newpage
\section{Data Preprocessing Flowchart}\label{appendixE}
\begin{figure*}[h]
\centerline{\includegraphics[width=0.9\textwidth]{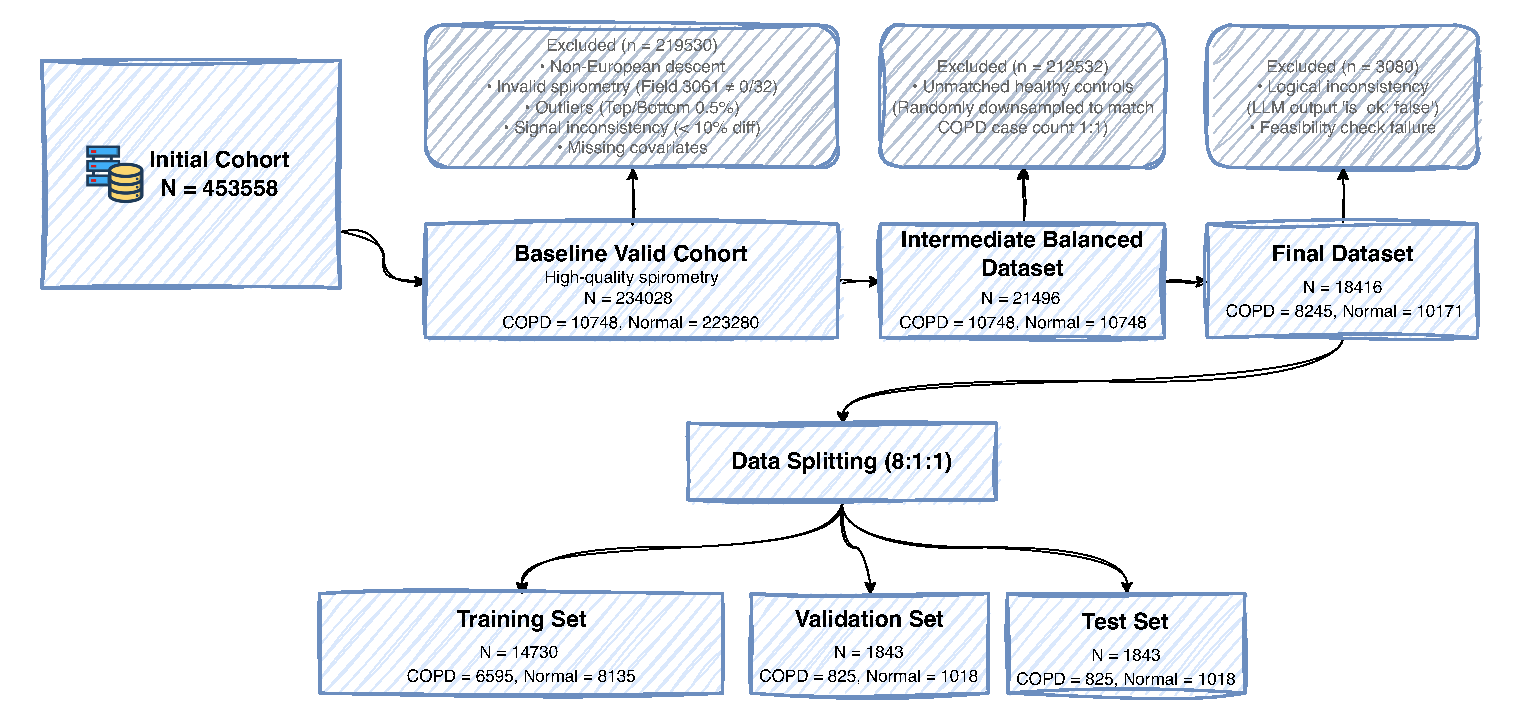}}
\end{figure*}

\section{Expert Evaluation Results}\label{appendixF}
\begin{figure*}[h]
\centerline{\includegraphics[width=1\textwidth]{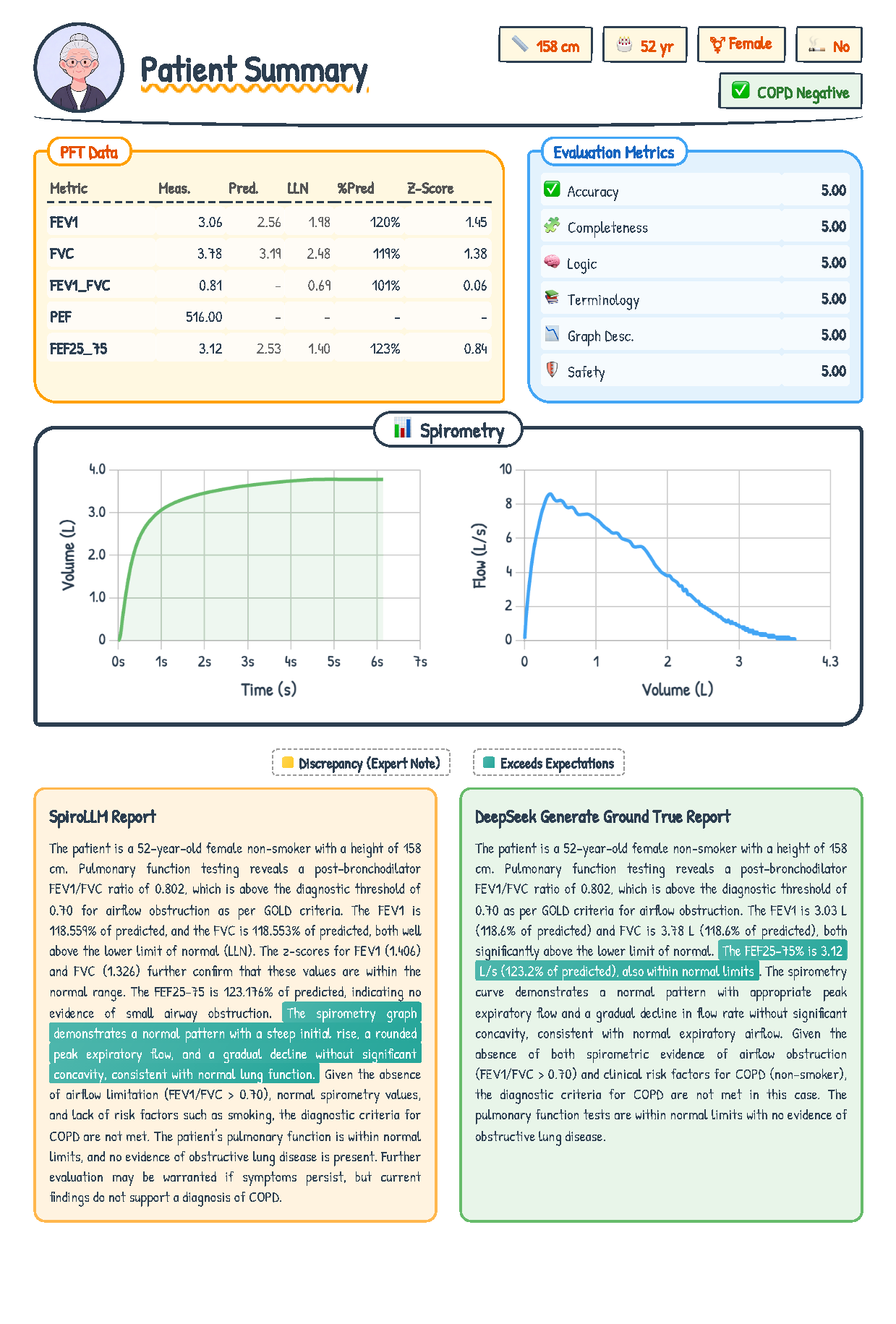}}
\end{figure*}

\begin{figure*}[h]
\centerline{\includegraphics[width=1\textwidth]{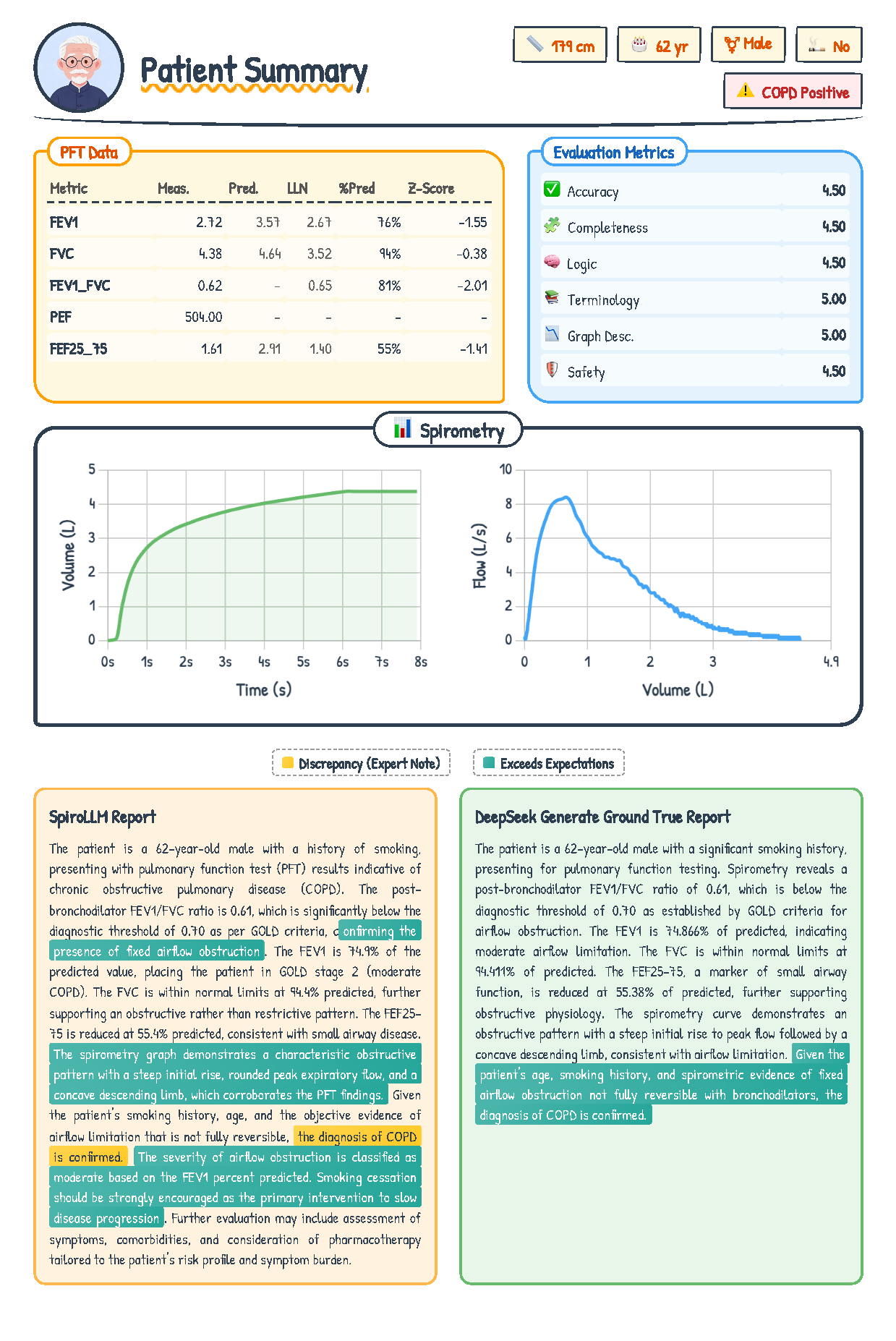}}
\end{figure*}

\begin{figure*}[h]
\centerline{\includegraphics[width=0.95\textwidth]{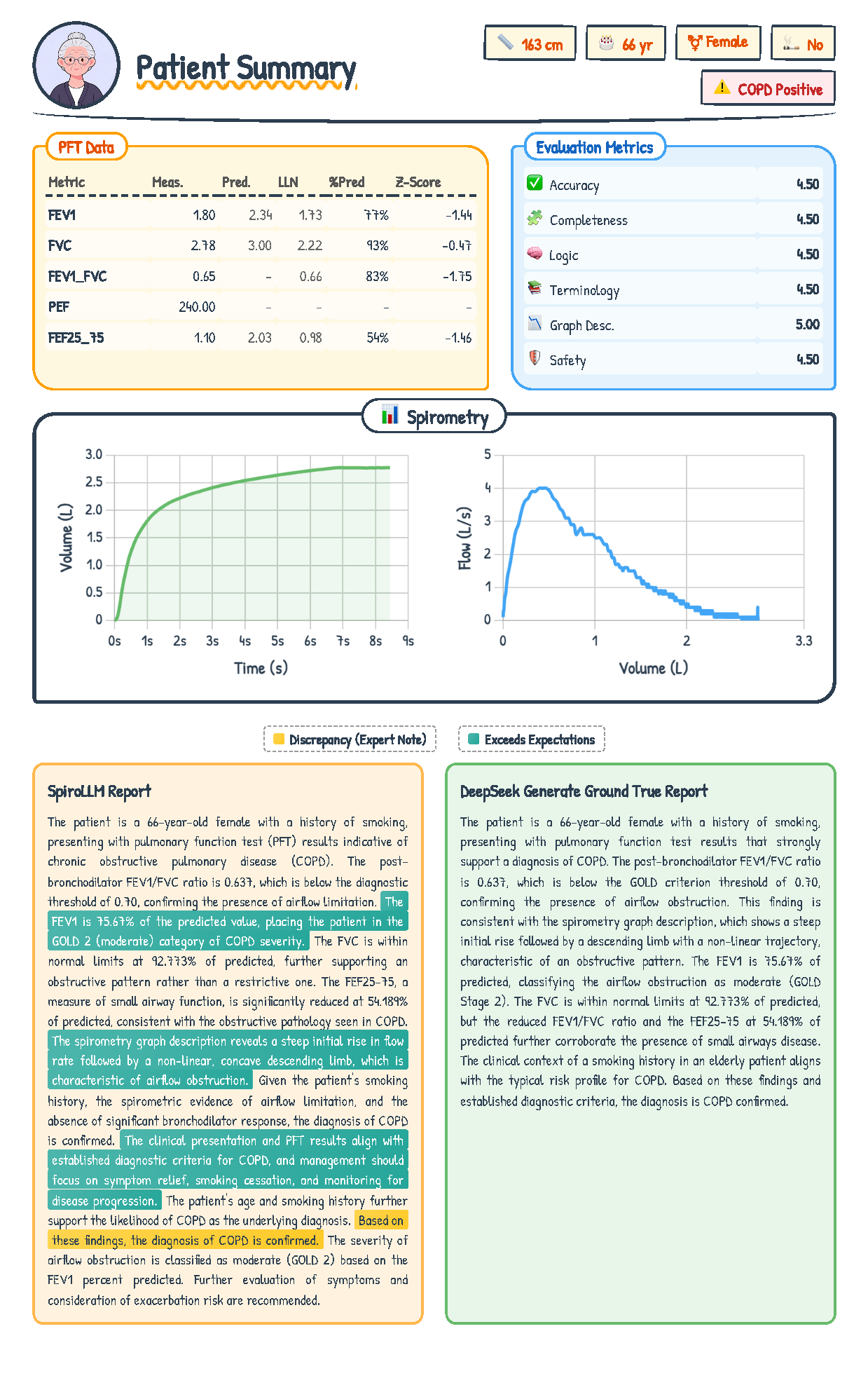}}
\end{figure*}

\begin{figure*}[h]
\centerline{\includegraphics[width=1\textwidth]{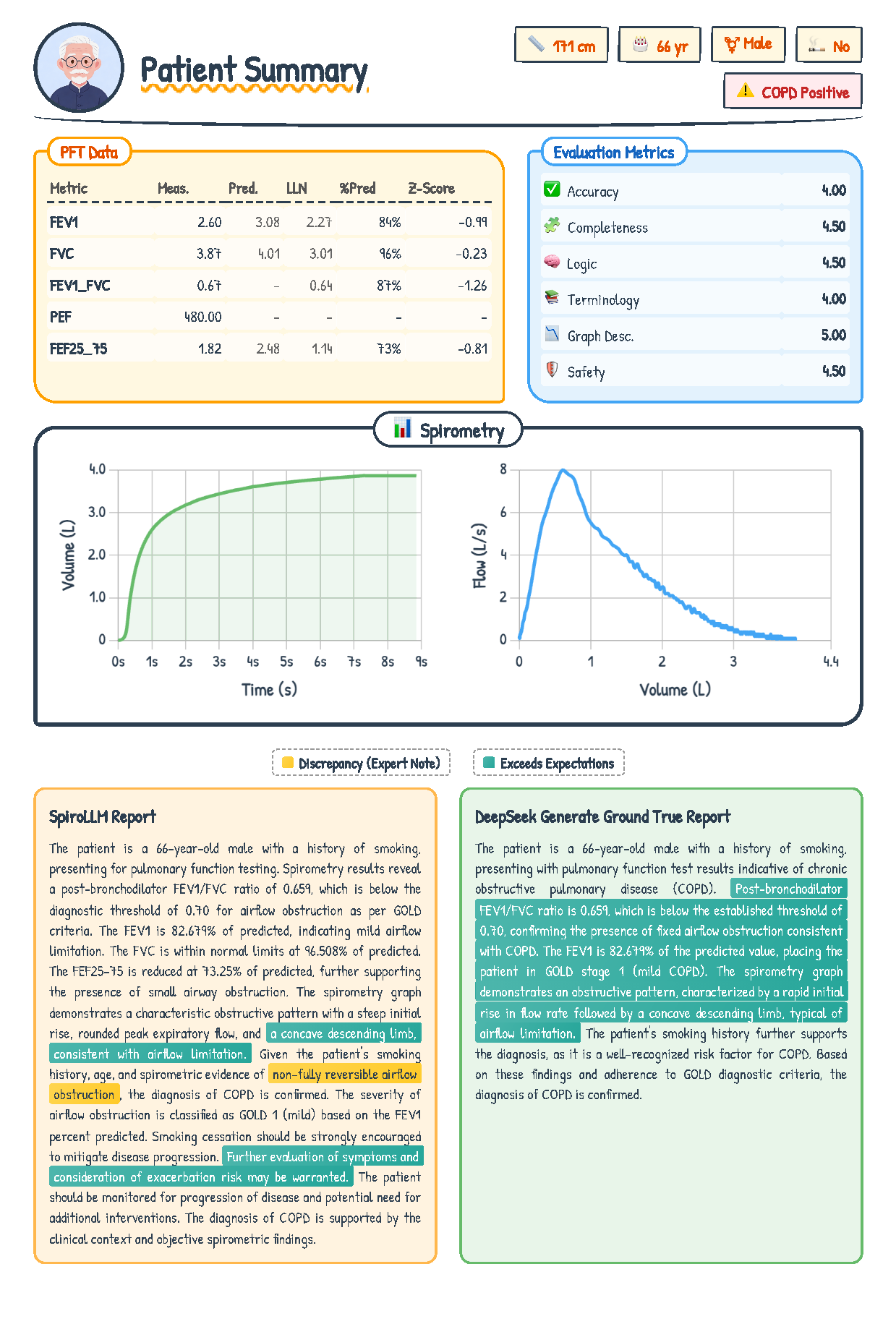}}
\end{figure*}

\end{appendices}

\end{document}